\documentclass[final,1p,times]{elsarticle}

\usepackage{amssymb}

\usepackage{adjustbox} 
\usepackage{multirow} 
\usepackage{subfigure}  
\usepackage{amsmath}    
\usepackage{stmaryrd}   
\usepackage{amsfonts}
\usepackage{tabularx}
\usepackage{capt-of}
\usepackage{amsfonts} 

\usepackage{svg} 

\usepackage{hyperref}

\usepackage{makecell} 

\usepackage{lipsum,graphicx,multicol} 

\usepackage{booktabs} 

\usepackage{subcaption}

\journal{arXiv}

\begin{document}

\begin{frontmatter}




\title{Graph Dual-stream Convolutional Attention Fusion for Precipitation Nowcasting}

\author{Lóránd Vatamány}
\author{Siamak Mehrkanoon\corref{cor1}}
\ead{s.mehrkanoon@uu.nl} 
\cortext[cor1]{Corresponding author}



\address{Department of Information and Computing Sciences, Utrecht University, Utrecht, The Netherlands}

\begin{abstract}

Accurate precipitation nowcasting is crucial for applications such as flood prediction, disaster management, agriculture optimization, and transportation management. While many studies have approached this task using sequence-to-sequence models, most focus on single regions, ignoring correlations between disjoint areas. We reformulate precipitation nowcasting as a spatiotemporal graph sequence problem. Specifically, we propose Graph Dual-stream Convolutional Attention Fusion, a novel extension of the graph attention network. Our model’s dual-stream design employs distinct attention mechanisms for spatial and temporal interactions, capturing their unique dynamics. A gated fusion module integrates both streams, leveraging spatial and temporal information for improved predictive accuracy. Additionally, our framework enhances graph attention by directly processing three-dimensional tensors within graph nodes, removing the need for reshaping. This capability enables handling complex, high-dimensional data and exploiting higher-order correlations between data dimensions. Depthwise-separable convolutions are also incorporated to refine local feature extraction and efficiently manage high-dimensional inputs. We evaluate our model using seven years of precipitation data from Copernicus Climate Change Services, covering Europe and neighboring regions. Experimental results demonstrate superior performance of our approach compared to other models. Moreover, visualizations of seasonal spatial and temporal attention scores provide insights into the most significant connections between regions and time steps.

\end{abstract}




\end{frontmatter}


\section{Introduction}
\label{sec:sample1}

Precipitation nowcasting involves forecasting the forthcoming intensity of rainfall typically on a timescale ranging from minutes to a few hours. Nowcasting can help the operations of several weather dependent sectors including energy management, retail, flood, traffic control and emergency services \cite{wilson2010nowcasting}. To serve these sectors effectively, the accuracy of nowcasting must extend across a range of spatial and temporal scales. Two primary approaches are commonly employed for precipitation nowcasting. The first one involves ensemble numerical weather prediction (NWP) systems, which rely on the physical properties of the atmosphere to generate multiple realistic precipitation forecasts. However, these methods are not suitable for short-term predictions due to their high computational expense, sensitivity to noise, and dependence on the initial conditions of the event \cite{heye2017precipitation}.
The second approach is optical flow methods, which derive velocity fields from consecutive images, are typically used as baseline predictions \cite{ayzel2019optical}. Despite being unsupervised and computationally efficient, optical flow techniques are limited by their simplistic assumptions and often fail to capture the nonlinear dynamics of precipitation events \cite{kumar2020convcast}.

In contrast to NWP models, data-driven approaches do not rely on the physical properties of the atmosphere. Instead, they utilize historical weather observations to train models capable of mapping input data to target outputs \cite{mehrkanoon2019deep}. Among these data-driven models, deep neural network architectures stand out, as they are trained in an end-to-end fashion and possess the ability to extract complex underlying patterns from data by incorporating multiple nonlinear layers. Recent advances in deep learning have showcased remarkable progress in the fields of weather element forecasting and related nowcasting tasks \cite{hess2022deep, han2022short, cho2020comparative, trebing2021smaat,fernandez2021broad, yang2022aa, fernandez2022deep, abdulla2022design, prado2024multivariable, ghimire2022hybrid, zhang2023skilful}. Similar approaches have also been successfully applied in fields like environmental risk assessment \cite{benmakhlouf2023landslide} and remote sensing \cite{dibs2023multi, dibs2023fusion}.

In particular, Convolutional Neural Networks (CNNs) based models have already demonstrated success in addressing the weather forecasting challenge \cite{fukuoka2018wind, trebing2020wind}. However, it's worth noting that CNN-based methods typically do not account for the spatial relationships between weather stations. In a prior study \cite{mehrkanoon2019deep}, the approach involved transforming historical data into a tensor format (comprising weather stations, weather variables, and time steps), which was subsequently fed into the model, with convolution operations applied across the data volume. Consequently, the neighborhood relationships between weather stations were primarily determined by their order in the dataset rather than explicitly considering their spatial proximity.

Weather patterns are inherently spatial, with various meteorological factors interacting across geographical regions. Therefor, graph neural networks (GNNs) based models that can generalize CNNs to work on graphs rather than on regular grids are among promising architecture for the weather elements nowcasting. In particular, GNNs can capture the intricate spatial dependencies by modeling data as a graph, where nodes represent locations or weather stations, and edges represent the connections between them. This enables GNN based models to account for the influence of neighboring regions on each other's weather conditions, making them effective at modeling spatial correlations. In addition, GNNs can also be extended to incorporate temporal information, allowing them to model how weather conditions change over time. This is crucial for short-term weather predictions and nowcasting. However, existing graph-based models often struggle with high-dimensional data at the nodes, leading to the need for reshaping, which can result in the loss of critical information, particularly when dealing with complex data structures like images.

Despite these advancements, there are still several challenges in precipitation nowcasting. As mentioned earlier, CNN models often fail to consider multiple regions simultaneously, which can lead to sub-optimal predictions when dealing with spatially diverse weather data. On the other hand, graph-based models often struggle with high-dimensional data at their nodes, which forces the reshaping of this data. In the case of images, this reshaping can lead to a loss of critical information.

In this paper we propose a novel Graph Dual-stream Convolutional Attention Fusion (GD-CAF)\footnote{https://github.com/wendig/GD-CAF} a novel architecture for improving weather nowcasting. The key contributions of our work are as follows:

\begin{enumerate}
    \item We introduce novel spatiotemporal convolutional attention and gated fusion modules, enhanced with depthwise-separable convolutional operations. This augmentation enables the model to effectively capture and leverage the inherent correlations and dependencies in spatiotemporal graph sequences, leading to improved nowcasting performance. Unlike many other graph-based models, such as \cite{zheng2020gman, velickovic2017graph}, which require one-dimensional node features, the GD-CAF model can directly analyze high-dimensional, tensorial node features within the spatiotemporal graph of precipitation maps. This versatility allows GD-CAF to handle richer and more complex data representations, thereby enhancing its predictive accuracy in weather nowcasting.
    \item We evaluate our model using precipitation data from the Copernicus Emergency Management Service (CEMS). We collected seven years of hourly precipitation maps (2016-2022) for the Euro-Asian region, focusing on 16 distinct areas represented as nodes in a graph. The model is trained on six years of data and tested on the final year. Our experiments examine various graph sizes and prediction horizons to assess the model's performance in nowcasting precipitation.
    \item We conduct a comparative analysis by benchmarking GD-CAF against Persistence, SmaAt-UNet \cite{trebing2021smaat} and RainNet \cite{ayzel2020rainnet} models, exploring various graph sizes and prediction horizons to evaluate their performance in precipitation nowcasting.
    \item We analyze the spatial and temporal attention mechanisms of our model using data from the test set. We provide a detailed examination of how attention is distributed across different regions and time steps, exploring spatial relationships and temporal dependencies. Our approach offers insights into the patterns and correlations captured by the model, enhancing our understanding of its capability to integrate and process complex spatial and temporal information.
\end{enumerate}

This paper is organized as follows. A brief overview of the related research works is given in Section \ref{sec:related:work}. Section \ref{sec:methods} introduces the proposed GD-CAF model. The experimental settings and description of the used datasets are given in Section \ref{sec:experiments}. The obtained results are discussed in Section \ref{sec:results:discussion} and the conclusion is drawn in Section \ref{sec:conclusion}.

\section{Related Work}
\label{sec:related:work}

\begin{figure*}
    \centering
    \includegraphics[width=\textwidth,keepaspectratio]{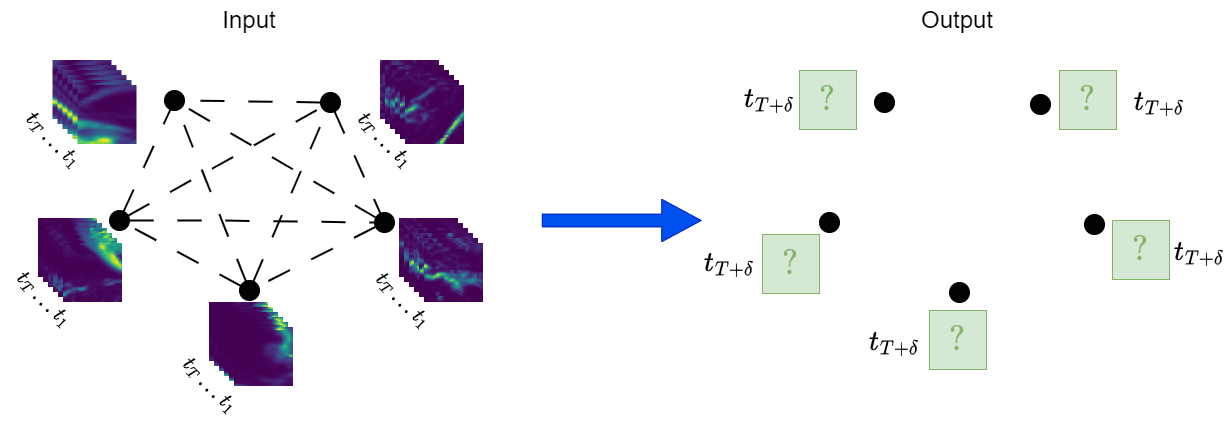}
    \caption{The input is a spatiotemporal graph sequence, and the output is a spatiotemporal graph.}
    \label{fig:graph_in_out_overview}
\end{figure*}

Weather element forecasting based on deep-learning architectures has recently gained a lot of attention due to the availability of large amount of weather data and the rapid advances in neural network techniques. The literature has already witnessed successful application of different architectures including Recurrent Neural Network (RNN)  \cite{cao2012forecasting}, Long short-term memory (LSTM) \cite{zaytar2016sequence}, Convolutional LSTM (ConvLSTM) \cite{shi2015convolutional}, Convolutional Neural Network (CNN), encoder-decoder \cite{larraondo2019data}, UNet \cite{agrawal2019machine} and graph neural networks \cite{simeunovic2021spatio} in weather forecasting and nowcasting related tasks.  

For instance, the authors in \cite{shi2015convolutional}, introduced a convolutional LSTM model to predict future rainfall intensity in Hong Kong over a relatively short period. In \cite{kumar2020convcast}, the authors proposed Convcast, an embedded convolutional LSTM-based architecture. Additionally, \cite{klein2015dynamic} introduced a dynamic convolutional layer for short-range weather prediction. Furthermore, in \cite{guastavino2022prediction}, a deep convolutional neural network is employed for predicting thunderstorms and heavy rains. A CNN-based wind speed prediction model that effectively captures spatiotemporal patterns in wind data using real weather datasets from Denmark and the Netherlands is presented in \cite{trebing2020wind}. The author in \cite{mehrkanoon2019deep} proposed different CNN architectures, including 1-D, 2-D, and 3-D convolutions, to accurately predict wind speed and temperature for few hours ahead.

The UNet architecture, initially a successful model primarily used in the field of medical image analysis, has found application in precipitation nowcasting as well, as demonstrated in the study by Lebedev et al. \cite{lebedev2019precipitation}. In a subsequent work by Trebing et al. \cite{trebing2021smaat}, a model known as SmaAt-UNet was introduced as an extension of the core UNet model. This extended model significantly reduces the number of parameters in the UNet without compromising its performance, as outlined in \cite{trebing2021smaat}. Furthermore, in another study by Fernandez et al. \cite{fernandez2021broad}, a modification called Broad-UNet was introduced, which enhances the UNet architecture by incorporating asymmetric parallel convolutions and the Atrous Spatial Pyramid Pooling (ASPP) module. 

Diffusion models are gaining traction in precipitation nowcasting due to their ability to produce high-quality, detailed predictions. For instance, Diffcast \cite{yu2024diffcast} uses residual diffusion to model precipitation dynamics, reducing blurriness and positional errors. Similarly, Prediff \cite{gao2024prediff} combines latent diffusion with knowledge alignment to produce probabilistic forecasts that respect physical constraints. Additionally, CasCast \cite{gong2024cascast} introduces a cascaded framework that combines deterministic and probabilistic modeling to improve predictions for complex and extreme precipitation events, achieving significant performance gains in high-resolution scenarios.

Although the previous works exploit spatio-temporal correlations, they do not fully leverage the spatial information from multiple weather stations (regions). Graph neural networks (GNNs) \cite{gori2005new} have recently attracted a lot of attention due to their expressive power and ability to infer information from complex data, such as brain signals, social network interactions, and weather prediction \cite{zhou2020graph, wu2020comprehensive, stanczyk2021deep, aykas2021multistream}. They propagate information through the graph nodes and edges, enabling the model to capture the underlying structure and dependencies within the graph. For instance, the authors in \cite{keisler2022forecasting} utilized a graph neural network for global weather forecasting, where the system learns to project the current 3D atmospheric state six hours ahead, yielding improved results. The authors in \cite{owerko2018predicting} used graph neural networks to predict power outages based on current weather conditions. 

One category of Graph Neural Networks (GNNs) includes Graph Convolution Networks (GCNs) as described in \cite{kipf2016semi}. GCNs extend the capabilities of Convolutional Neural Networks (CNNs) to operate on graphs rather than regular grids. They are particularly adept at integrating neighbor relationships, often through the adjacency matrix of a graph. In the study by Stanczyk et al. \cite{stanczyk2021deep}, the authors applied graph convolutional networks (GCNs) to tackle the challenge of wind speed prediction using data from multiple weather stations. Their model outperformed existing baseline methods when tested with real datasets from weather stations in Denmark and the Netherlands. Similarly, in \cite{simeunovic2021spatio}, the authors introduced GCLSTM and GCTrafo, graph convolutional models used to address solar power generation forecasting from multi-site photovoltaic production data represented as signals on a graph. These models, solely reliant on production data, surpassed existing multi-site forecasting methods, especially for a six-hour prediction horizon.

Another category of GNNs comprises Graph Attention Networks (GAT) \cite{velivckovic2017graph}, which are designed to work with graph data and leverage attention mechanisms. An extension of GAT, known as the Heterogeneous Graph Attention Network, was introduced by Wang et al. in \cite{wang2019heterogeneous}. This approach handles the complexities associated with heterogeneous graphs containing different types of nodes and links by incorporating node-level and semantic-level attentions. Aykas et al. extended GAT further in \cite{aykas2021multistream}, introducing Multistream Graph Attention Networks. This model incorporates a learnable adjacency matrix and a novel attention mechanism, which they applied to predict wind speeds for multiple cities. In \cite{mi2020hierarchical}, a Hierarchical Graph Attention Network was proposed by the authors to capture dependencies at both the object-level and triplet-level, allowing the model to represent interactions between objects and dependencies among relation triplets. For spatial-temporal analysis, Spatial-Temporal Graph Attention Networks were introduced in \cite{zhang2019spatial}, where graph attention mechanisms were used to capture spatial dependencies among road segments, and LSTM networks were employed to extract temporal features.

In what follows, we propose a novel Graph Dual-stream Convolutional Attention Fusion model, which enhances traditional graph attention networks by employing distinct attention mechanisms for both spatial and temporal interactions. These mechanisms are integrated through a gated fusion module, while depthwise-separable convolutions are employed to more effectively capture complex, high-dimensional data. We evaluate the model on precipitation data from multiple regions, showcasing its enhanced performance and interpretability through attention-based visualizations.

\section{Methods}
\label{sec:methods}

\subsection{Graph precipitation maps}

\fontdimen16\textfont2=3pt
\fontdimen17\textfont2=3pt

We denote a network of precipitation maps as a fully connected graph $G = (V, E)$. Here, V is a set of $N = |V|$ vertices (nodes), and $E$ represents the edges connecting them. Each node holds historical observations from a particular region, see the input graph in Fig. \ref{fig:graph_in_out_overview}.
Given the precipitation maps with $H$ (height) and $W$ (width) dimensions at $N$ vertices over $T$ time steps, we denote one sample as $\omega$ in the dataset $\mathcal{D}$ with input $X$ and output $Y$ as follows: 
\begin{equation}
\omega = \biggl\{X, Y \biggr\} = \biggl\{x_{v_i, t_j}, y_{v_i,\,t_\Delta}\biggr\}_{i=1,\,j=1}^{N, T},
\end{equation}
where $x_{v_i,t_j} \in \mathbb{R}^{H \times W}$ is a precipitation map at time step $t_j$ for node $v_i$. The target $y_{v_i,\,t_\delta} \in \mathbb{R}^{H \times W}$ is the precipitation map in the future time step ($t_{\Delta=T+\delta}$) for node $v_i$. Given the input graph, the goal is to predict the precipitation maps of all nodes for a single time step into the future as illustrated in Fig. \ref{fig:graph_in_out_overview}.

\subsection{Proposed Model}

Here, we introduce the Graph Dual-stream Convolutional Attention Fusion (GD-CAF) model that leverages higher-order correlations among the node dimensions of the historical spatiotemporal graph of precipitation maps and nowcast precipitation for a time step ahead at various spatial locations. GD-CAF is composed of spatio-temporal convolutional attention and gated fusion modules, both of which incorporate depthwise-separable convolutional operations. Unlike other competing graph-based models such as \cite{zheng2020gman, velickovic2017graph, wang2019heterogeneous} that necessitate low-dimensional node input representation, our proposed model can directly process high-dimensional nodes. As a result, the fundamental structure of the graph information remains unchanged.

\begin{figure*}
    \centering
    \includegraphics[width=0.86\textwidth,keepaspectratio]{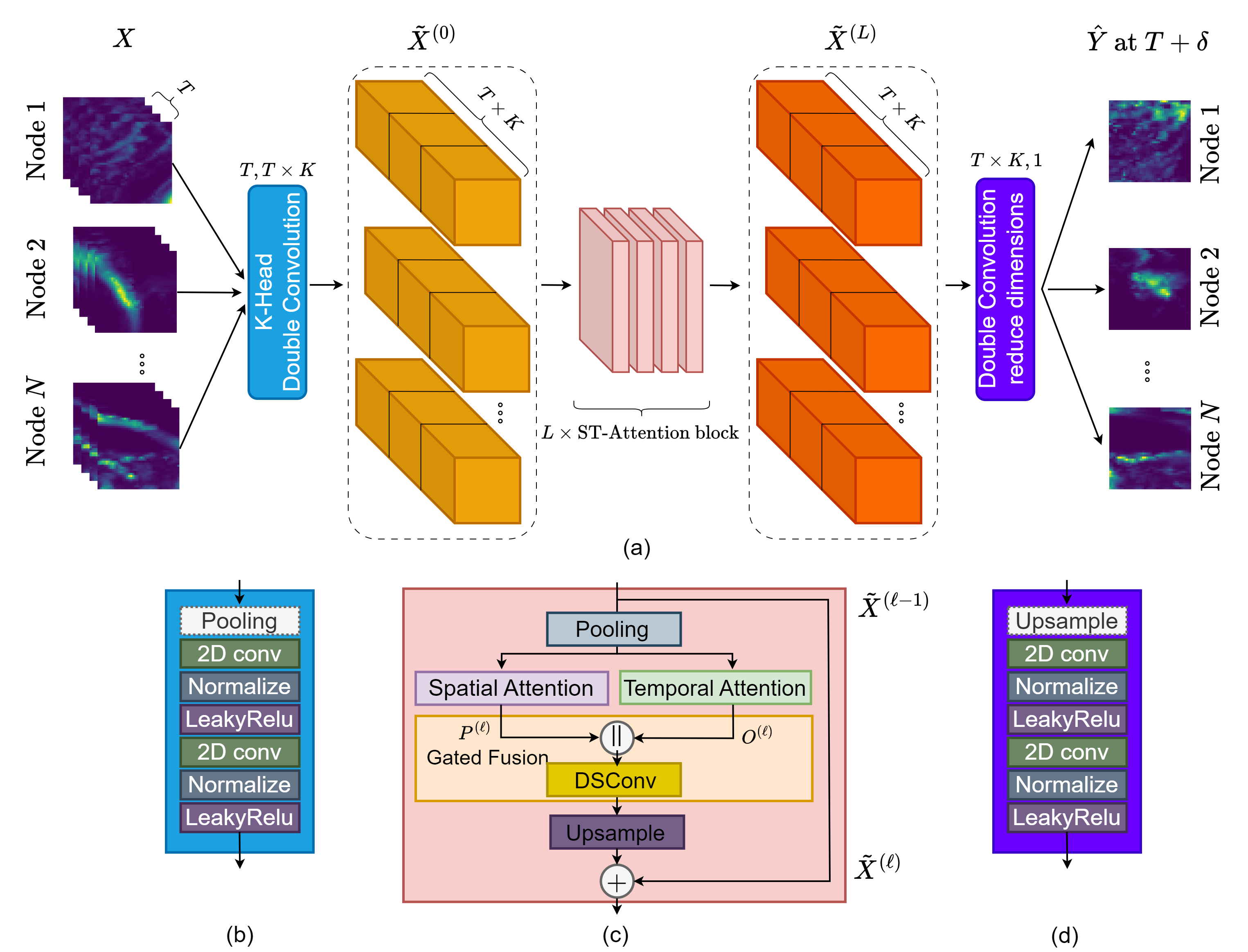}
    \caption{(a) GD-CAF architecture overview (b) Double convolution block with pooling (c)  ST-Attention block in the $\ell$-th block. (d) Double convolution block with upsampling. The numbers above the double convolutional blocks indicate the number of input and output channels respectively.}
    \label{fig:conv_gman_overview}
\end{figure*}

An overview of the proposed GD-CAF model is illustrated in Fig. \ref{fig:conv_gman_overview} (a). The input to the model, i.e. $X = \left\{x_{v_i, t_j} \in \mathbb{R}^{H \times W}\right\}_{i=1, j=1}^{N, T}$, consists of historical observations at $N$ nodes for $T$ time steps. The input $X$ is initially transformed into $\tilde{X}^{(0)}$ using a double convolutional operation, during which the temporal depth increases $K$ times, corresponding to the number of heads used in the attention mechanisms. Therefore, $\tilde{X}^{(0)}$ is defined as $\left\{\tilde{x}_{v_i, t_j} \in \mathbb{R}^{H \times W}\right\}_{i=1, j=1}^{N, T \times K}$. Next, $\tilde{X}^{(0)}$ is passed through a sequence of $L$ ST-Attention blocks, producing the output $\tilde{X}^{(L)}$ at the $L$-th block, with the same dimension as $\tilde{X}^{(0)}$. The depth dimension ($T \times K$) of the output of the ST-Attention blocks are then reduced using a double convolutional operation to obtain a single time step for all nodes, represented as $\hat{Y} = \left\{y_{v_i, t_\Delta} \in \mathbb{R}^{H \times W}\right\}_{i=1}^{N}$. The proposed GD-CAF is trained in an end-to-end fashion by minimizing the mean squared error (MSE) between the predicted and the ground-truth precipitation maps. It's worth highlighting that when striving to establish a universal representation across all nodes, shared filters are utilized for every graph node throughout all convolutional operations.

\subsubsection{ST-Attention Block}

As illustrated in Fig. \ref{fig:conv_gman_overview} (c), the ST-Attention block consists of pooling, upsampling as well as spatial and temporal attention modules which are combined through a gated fusion. Furthermore, multi head attention is used to stabilize the learning process for spatial and temporal attention. Therefore, the attentions are computed $K$ times with different learnable nonlinear projections. As opposed to graph attention based models \cite{zheng2020gman, velickovic2017graph, wang2019heterogeneous} that use fully-connected layers for computing the queries and keys, here we introduce convolutional operations as nonlinear projections. This enables us to learn directly from high-dimensional node representation without the need of flattening them, therefore the structure of data remains unchanged. In addition the total number of trainable parameters are also reduced. In particular, we use depthwise-separable convolution \cite{chollet2017xception} where convolutions are applied separately on individual input channels and then combined. This results in a substantial decrease in the number of trainable parameters when compared to standard convolution, leading to a more efficient model with lower computational complexity and memory demands.

In situations where the quantity of nodes greatly surpasses the number of features within each node, numerous approaches have been suggested in the literature to mitigate computational complexity. These techniques include node grouping and contextual attention approaches, as discussed in \cite{kitaev2020reformer, wang2020linformer, shen2021efficient}. In spatiotemporal graphs, where the number of features exceeds the number of nodes, reducing computational complexity can for instance be achieved through pooling operations. Pooling involves downsampling or aggregating information thereby reducing spatial dimension and accelerating overall computation. 

\subsubsection{Spatial attention}

\begin{figure*}
    \begin{minipage}[c]{0.5\textwidth}
        \centering
        \includegraphics[width=\linewidth]{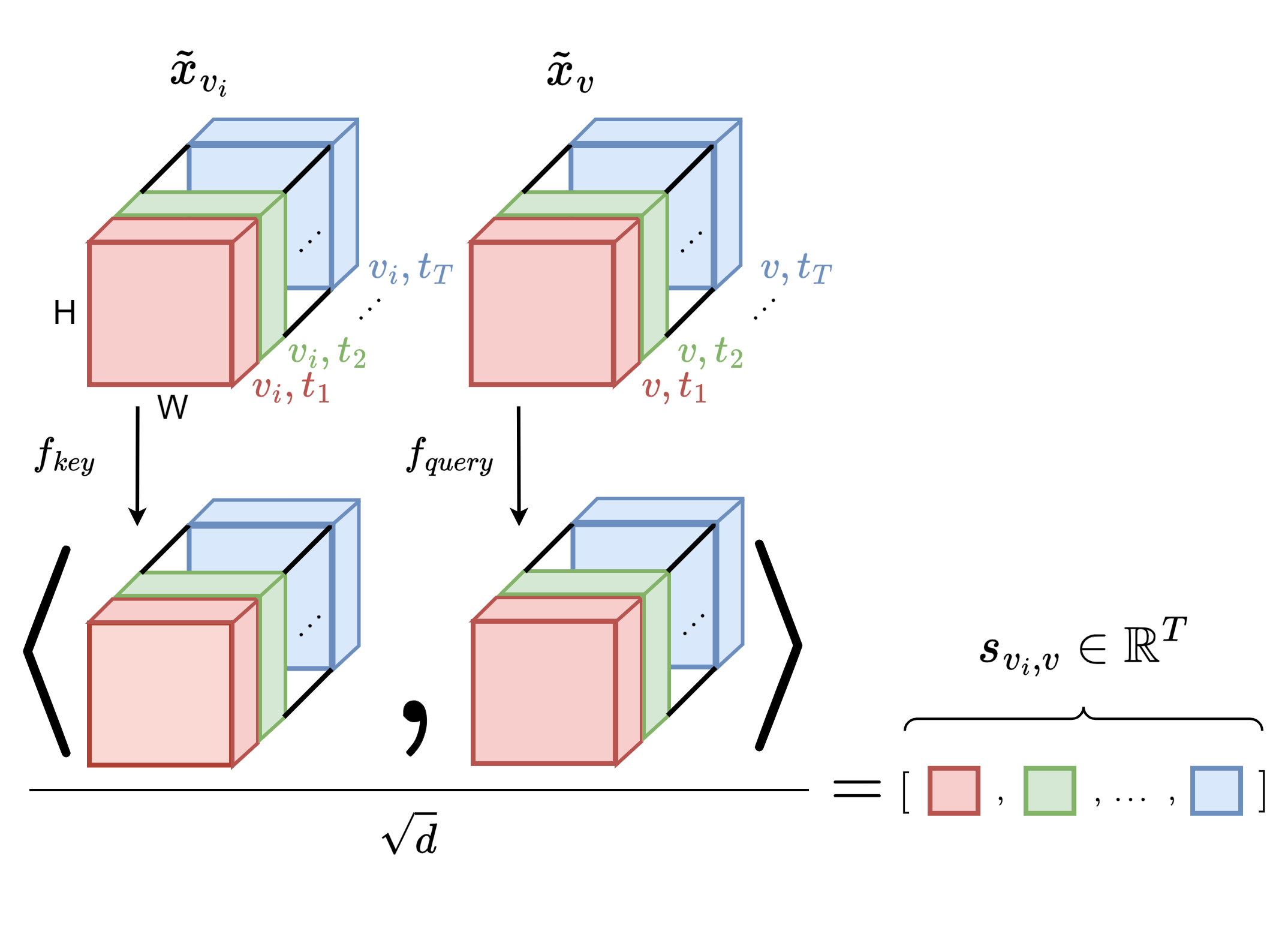}
        \caption*{(a)}
    \end{minipage}\hfill
    \begin{minipage}[c]{0.5\textwidth}
        \centering
        \includegraphics[width=\linewidth]{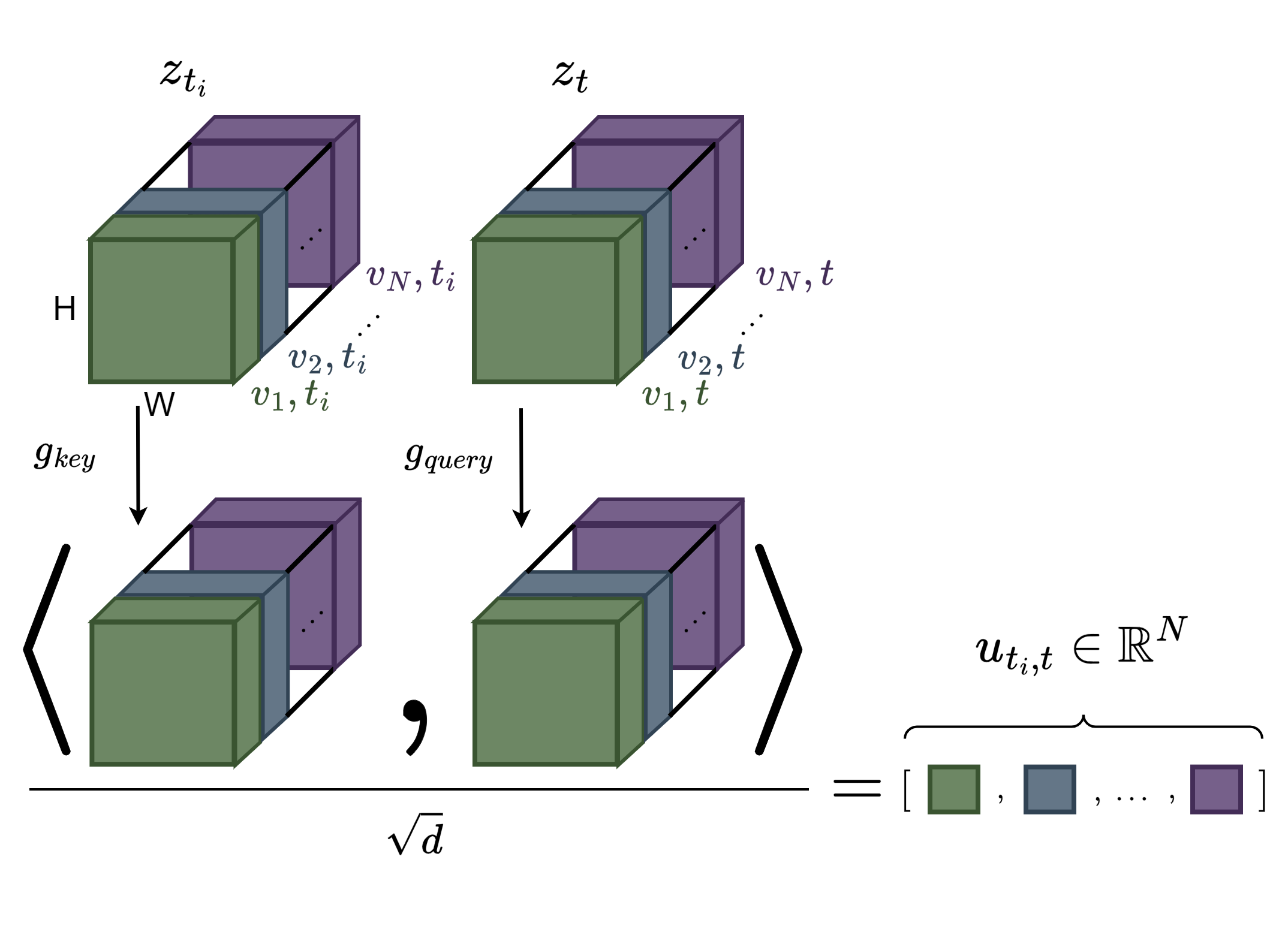}
        \caption*{(b)}
    \end{minipage}
    \caption{Spatial and temporal attention on 3D tensors with only one attention head. (a) Spatial attention is calculated between different nodes, at the same time step. (b) Temporal attention is calculated within one node, but between different time steps.}
    \label{fig:spatial_temporal_with_tensors}
\end{figure*}

Let us denote the input representation corresponding to the $k$-th head in the $\ell$-th ST-Attention block at node $v_i$ as follows:

\begin{equation}
    \tilde{x}^{(\ell-1),(k)}_{v_i}= [\tilde{x}^{(\ell-1)}_{v_i,\,t_{_{kT+1}}}, \tilde{x}^{(\ell-1)}_{v_i,\,t_{_{kT+2}}}, \ldots, \tilde{x}^{(\ell-1)}_{v_i,\,t_{_{(k+1)T}}}] \in \mathbb{R}^{T \times H \times W},
\end{equation}

\text{and}
\[
\left\{
\begin{array}{l}
i: 1 \leq i \leq N \textrm{ is the node index}\\
k: 0 \leq k \leq K-1 \textrm{ is the head index}\\
\ell: 1 \leq \ell \leq L \textrm{ is index of the ST-Attention block.}
\end{array}
\right.
\]

Here, $\tilde{x}^{(\ell-1)}_{v_i,\,t_{_{kT+j}}} \in \mathbb{R}^{H \times W}$ is the input representation of the perception map of node $v_i$ for the $k$-th head at time step $t_{kT+j}$ in the $\ell$-th ST-Attention block.

The precipitation in a particular area is influenced by precipitation in other areas to varying degrees. To model this highly dynamic relationship, we extend the Scaled Dot-Product Attention \cite{vaswani2017attention} to dynamically assign weights to tensorial node representation, at each time steps, also illustrated in Fig. \ref{fig:spatial_temporal_with_tensors} (a). This relationship between tensorial node $v_i$ and $v$ is expressed as follows: 
\begin{equation}
    s_{v_i, v}^{(k)} = \frac{\langle f_{query}^{(k)} * \tilde{x}_{v_i}^{(\ell-1), (k)}, f_{key}^{(k)} * \tilde{x}_{v}^{(\ell-1), (k)}\rangle }{\sqrt{d}} \in \mathbb{R}^{T}.
\end{equation}
Here ‘$*$’ represents a double depthwise-separable convolution, consisting of the following sequential operations: depthwise convolution, pointwise convolution, normalization and a $\textrm{ReLU}$ activation function. $\langle \cdot, \cdot \rangle$ is the inner product operator, and $d$ is the size of a single precipitation map, which is used as a scaling factor. The relevance $s_{v_i, v}^{(k)}$ is a $T$ dimensional vector that represents the relationship between tensorial node $v$ and node $v_i$ in the $k$-th head. The attention scores are computed as follows:

\begin{equation}
    \alpha_{v_i, v}^{(k)} = \frac{exp \Big( \textrm{LeakyReLu}(s_{v_i, v}^{(k)}) \Big) }{\sum_{v_m \in V} exp \Big( \textrm{LeakyReLu}(s_{v_i, v_m}^{(k)}) \Big)} \in \mathbb{R}^{T},
\end{equation}
where $\alpha_{v_i, v}^{(k)}$ is the attention score between tensorial node $v_i$ and $v$ in the $k$-th head. By normalizing we ensure that for each node $\sum_{v \in V} \alpha_{v_i, v}^{(k)} = 1$. The output of the spatial attention module for node $v_i$ in $\ell$-th ST-Attention block is then computed as follows:

\begin{equation} \label{eq:spatial:output}
    p_{v_i}^{(\ell)} = \bigparallel_{k=1}^{K} \bigg( \sum_{v \in V} \alpha_{v_i, v}^{(k)} \Big( f_{value}^{(k)} * \tilde{x}_{v_i}^{(\ell-1), (k)} \Big) \bigg) \in \mathbb{R}^{T \times H \times W},
\end{equation}
where $\parallel$ is the concatenation operation and ‘$*$’, defined as previously, is applied to the transformed node representation $\tilde{x}_{v_i}^{(\ell-1), (k)}$ with learnable parameters $f_{value}^{(k)}$. Next, the obtained output of Eq. (\ref{eq:spatial:output}) is fed into a double convolutional operation to further enhance the feature representation by capturing more local patterns and structures.

\subsubsection{Temporal attention}

Precipitation in one area is influenced by the past values in the same area. To model this highly dynamic relationship, we extend the Scaled Dot-Product Attention \cite{vaswani2017attention}, also illustrated in Fig. \ref{fig:spatial_temporal_with_tensors} (b). Let us denote the input representation of all the nodes corresponding to the $k$-th head in the $\ell$-th ST-Attention block at time step $t_i$ as follows:

\begin{equation}
    z^{(\ell-1), (k)}_{t_i}= [\tilde{x}^{(\ell-1), (k)}_{v_1,\,t_{_{i}}}, \tilde{x}^{(\ell-1), (k)}_{v_2,\,t_{_{i}}}, \ldots, \tilde{x}^{(\ell-1), (k)}_{v_N,\,t_{_{i}}}] \in \mathbb{R}^{N \times H \times W},
\end{equation}

where
\[
\left\{
\begin{array}{l}
i: 1 \leq i \leq T \textrm{ is the time step},\\
k: 0 \leq k \leq K-1 \textrm{ is the head index},\\
\ell: 1 \leq \ell \leq L \textrm{ index of the ST-Attention block.}
\end{array}
\right.
\]

Here, $\tilde{x}^{(\ell-1),(k)}_{v,\,t_{i}} \in \mathbb{R}^{H \times W}$ is the input representation of the perception map, corresponding to the $k$-th head, of node $v$ at time step $t_{i}$ in the $\ell$-th ST-Attention block. The relevance between all the nodes at time step $t_i$, and $t$ in the $k$-th head can now be expressed as follows:
\begin{equation}
u_{t_i,\,t}^{(k)} = \frac{\langle g_{query}^{(k)} * z_{t_i}^{(\ell-1), (k)}\, , \, g_{key}^{(k)} * z_{t}^{(\ell-1), (k)}\rangle}{\sqrt{d}} \in \mathbb{R}^{N}.
\end{equation}
Here, the relevance $u_{t_i, t}^{(k)}$ is an $N$ dimensional vector. The symbol ‘$*$’ is defined as previously, and it is now applied using learnable parameters associated with $g_{query}^{(k)}$, and $g_{key}^{(k)}$.

$\langle \cdot, \cdot \rangle$ is defined as previously, and $d$ is the size of a single precipitation map, serving as a scaling factor. Subsequently, the attention scores are obtained as follows:
\begin{equation}
\beta_{t_i,\,t}^{(k)} = \frac{exp \Big( \textrm{LeakyReLu}(u_{t_i,\, t}^{(k)}) \Big) }{\sum_{t_m \in T} exp \Big(\textrm{LeakyReLu}(u_{t_i,\, t_m}^{(k)})\Big)} \in \mathbb{R}^{N},
\end{equation}
where the attention score $\beta_{t_i, t}^{(k)}$ is an $N$ dimensional vector in the $k$-th head, indicating the significance between time step $t$ and $t_i$. The output of the temporal attention module in the $\ell$-th block is then computed as follows: 

\begin{equation} \label{eq:temporal:output}
o_{t_i}^{(\ell)} = \bigparallel_{k=1}^{K} \bigg( \sum_{t \in T} \beta_{t_i,\,t}^{(k)} \Big( g_{value}^{(k)} * z_{t_i}^{(\ell-1), (k)} \Big) \bigg) \in \mathbb{R}^{N \times H \times W}.
\end{equation}
Here, $o_{t_i}^{(\ell)}$ is the output representation of the temporal attention in the $\ell$-th ST-Attention block, and ‘$*$’ is defined as previously and applied to $\tilde{z}_{t_i}^{(\ell-1), (k)}$ using learnable parameters associated with $g_{value}^{(k)}$. Next, in order to enrich the learned representations, the output of Eq. (\ref{eq:temporal:output}) is fed into a double convolutional operation. 

The above equations show how temporal attention in GD-CAF captures the correlations between different time steps for a given node. The convolution operation extracts local information from the node tensors, and the attention mechanism assigns weights to each time step based on their relevance to the target time step.

\subsubsection{Gated Fusion}

We use gated fusion to combine the obtained representations of spatial and temporal attention modules.
To this end, in the $\ell$-th ST-attention block, the outputs $P^{(\ell)}$ and $O^{(\ell)}$ are first concatenated followed by a double depthwise-separable convolution operation to create a new representation $\tilde{G}^{(\ell)}$ as follows:
\begin{equation}
    \tilde{G}^{(\ell)} = h_{gated} * \big( P^{(\ell)} \parallel O^{(\ell)} \big).
\end{equation}
Here, ‘$*$’ is defined as previously with learnable parameters $h_{gated}$, and concatenation is performed along the temporal depth. Gated fusion controls the flow between spatial and temporal attention at each node and time step. The output of the gated fusion is upsampled, in case pooling is used. Then the output of the $(\ell)$-th ST-Attention block is obtained as follows (see Fig. \ref{fig:conv_gman_overview}(c)):
\begin{equation}
    \tilde{X}^{(\ell)} = \tilde{X}^{(\ell-1)} + \textrm{upsample}(G^{(\ell)}).
\end{equation}

\subsection{Training}

\begin{figure}
    \centering
    \includegraphics[width=0.7\textwidth]{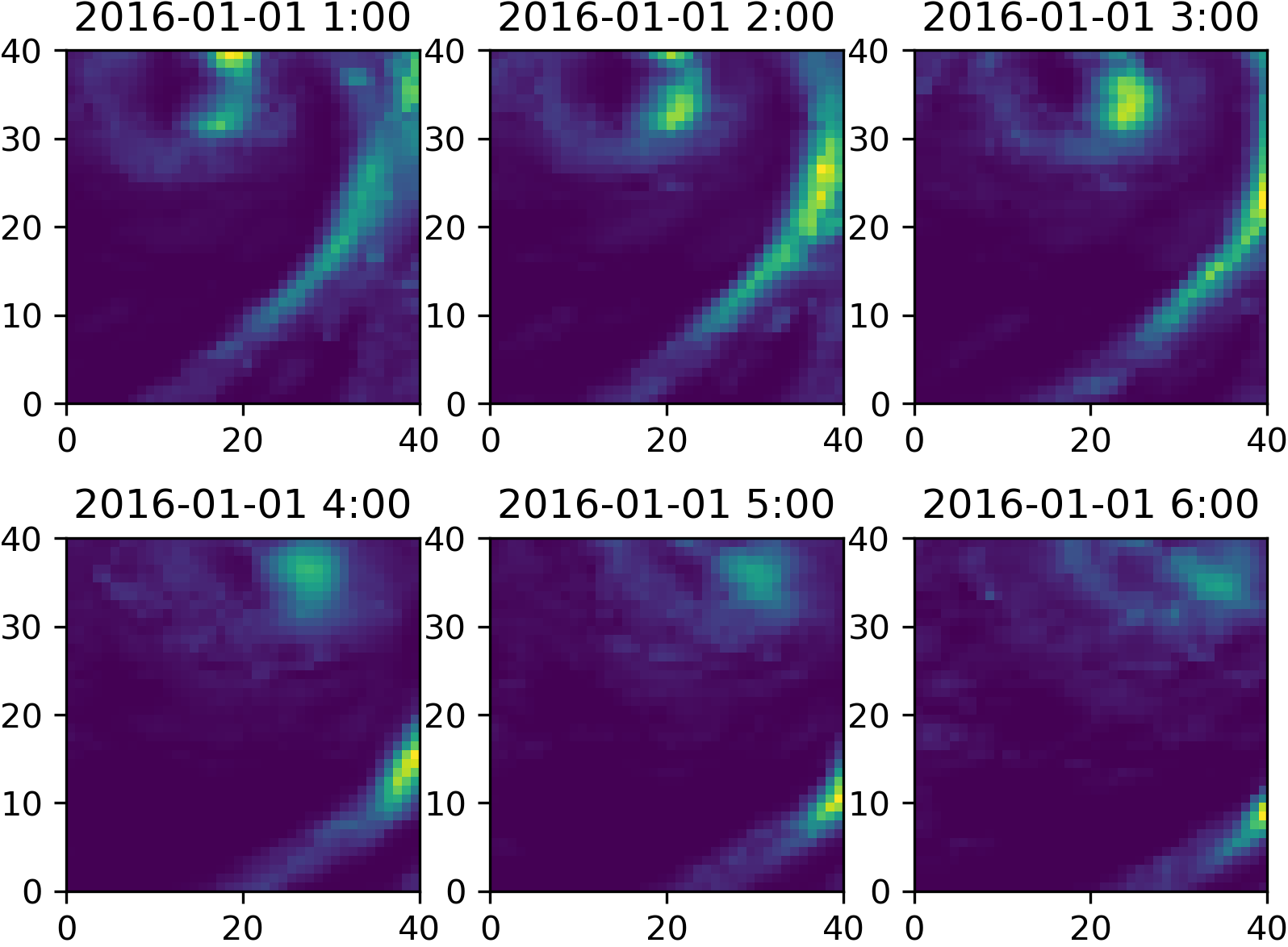}
    \caption{Precipitation data captured from the $4$-th region over a 6-hour time interval.}
    \label{fig:rain_seqience_0_6}
\end{figure}

All models are trained using PyTorch Lightning \cite{Falcon_PyTorch_Lightning_2019} framework for a maximum of 150 epochs. However, an early stopping criterion that keeps track of the validation loss is used. The training stops when the validation loss no longer decreases in the last 15 epochs. This criterion was met in all training iterations and the maximum of 100 epochs was never reached. Additionally, we used a learning rate scheduler that reduced the learning rate to a tenth of the previous learning rate when the validation loss did not improve for four consecutive epochs. The initial learning rate was set to 0.001 and we used the Adam optimizer \cite{kingma2014adam} with default values. The training was done on Google Colab Notebook \cite{bisong2019google} on a single Nvidia Tesla T4 graphics card that had 16 GB of memory.

We configured the hyperparameters of our model based on our empirical observations and also took into account that the compared models should have similar computational times. As a result, we opted for two ST-Attention blocks ($L=2$), each with four attention heads ($K=4$). For both GD-CAF and SmaAt-UNet, we used two kernels per channel within the depthwise-separable convolutional operations. The persistence model uses the most recent available observation for each region and treats it as the model prediction. An example of the nowcastings obtained by our proposed GD-CAF model is shown in Fig. \ref{fig:x_y_pred_18}.

\subsection{Model evaluation}
\label{sec:model_evaluation}

In order to evaluate the performance of our proposed model as well as the other examined models, we use the same metrics that are used in \cite{trebing2021smaat}.
Our main metric, the loss function used in this study is the mean squared error (MSE) between the predicted and the ground truth precipitation maps. 

\begin{equation}
   MSE = \frac{1}{n} \sum_{i=1}^{n} (Y_i - \hat{Y}_i)^{2},
\end{equation}
where $n$ is the number of samples, $Y_i$ is the ground truth for the $i$-th sample, which contains $N$ nodes, and $\hat{Y}_i$ is the predicted value for the $i$-th sample, also containing $N$ nodes.

Following the lines of \cite{trebing2021smaat},  in addition to the MSE, we also compute other metrices such as Precision, Recall (probability of detection), Accuracy and F1-score, critical success index (CSI), false alarm rate (FAR) and Heidke Skill Score (HSS). Similar to  \cite{trebing2021smaat}, these scores are calculated for rainfall bigger than a threshold of $0.5 mm/h$. To do this, we convert each pixel of the predicted output and target images to a boolean mask using this threshold. 
From this, one can calculate the true positives (TP) (prediction = 1, target = 1), false positives (FP) (prediction = 1, target = 0), true negatives (TN) (prediction = 0, target = 0) and false negatives (FN) (prediction = 0, target = 1). Subsequently, the CSI, FAR and HSS metrics can be computed as follows:

\begin{equation}
   CSI=\frac{TP}{TP + FP + FN},
\end{equation}

\begin{equation}
FAR=\frac{FP}{TP + FP},
\end{equation}

\begin{equation}
\begin{aligned}
HSS=\frac{2(TP \times TN) - 2(FP \times FN)}{(TP+FN)(FN+TN) + (TP + FP)(FP + TN)}.
\end{aligned}
\end{equation}

\section{Experiments}
\label{sec:experiments}

\begin{table*}[ht]
    \centering
    \renewcommand{\tabcolsep}{3pt}
    \begin{tabularx}{\textwidth}{Xcc}
        \begin{minipage}[c]{0.5\textwidth}
            \centering
            \includegraphics[width=0.9\linewidth]{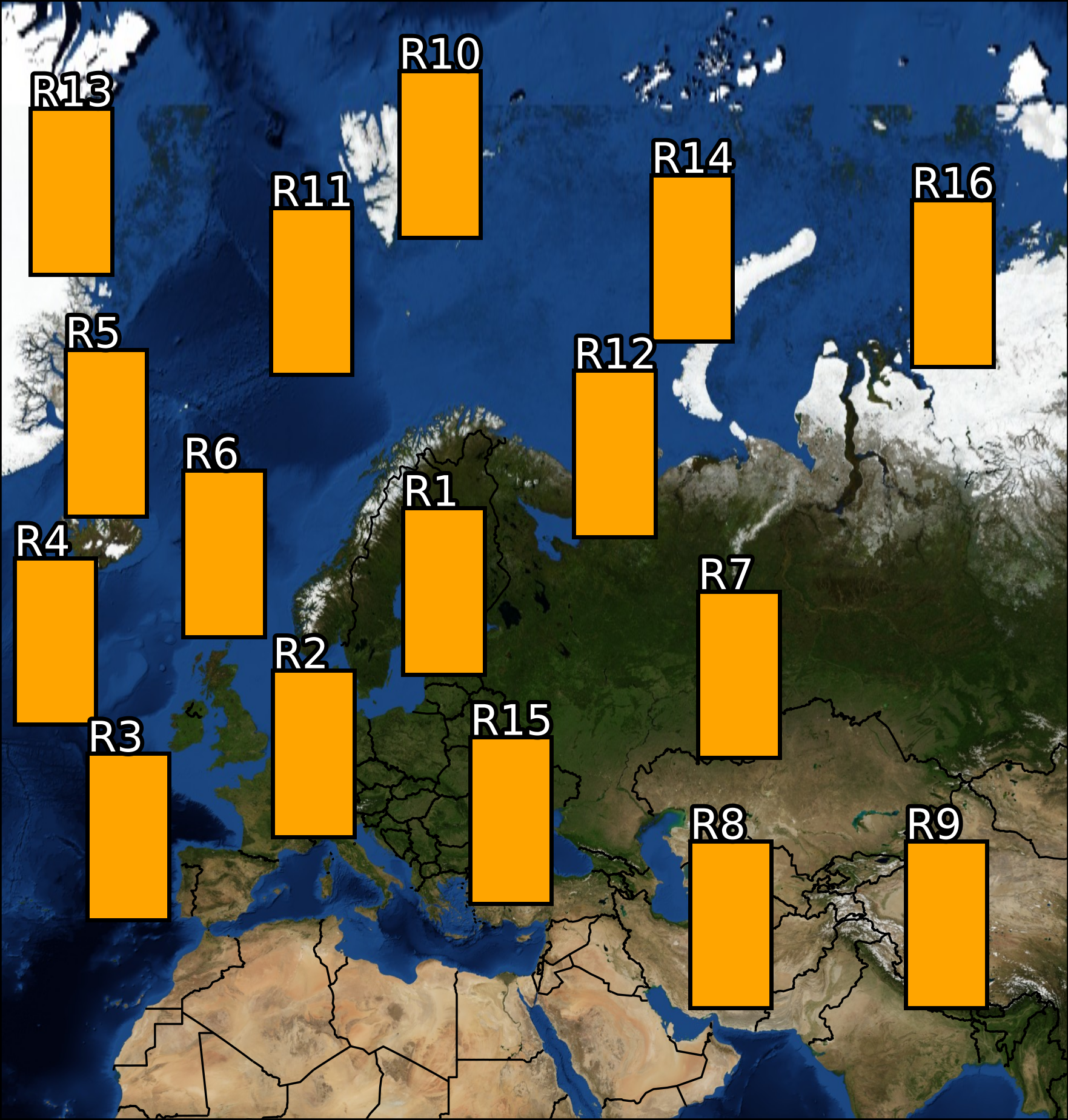}
            \captionof{figure}{Randomly placed areas on the map containing local precipitation maps.} 
            \label{fig:map_to_graph}
        \end{minipage}
        &  
        \begin{minipage}[c]{0.5\textwidth}
            \captionof{table}{16 selected regions, and their coordinates. LU means the left upper corner of the bounding box.}
            \label{Tab:region_coords}
            \centering
            \begin{tabular}{ c|cc }
                 \toprule
                Region id & LU* latitude & LU* longitude\\
                \hline
                1& 66.58 & 18.41\\
                2&57.31 & 2.44\\
                3&51.37 & -20.27\\
                4&64.03 &-29.25\\
                5&73.19 &-23.02\\
                6&68.34 &-8.54\\
                7&62.18 & 54.59\\
                8&44.19 & 53.59\\
                9&44.19 & 80.04\\
                10&80.70 & 17.91\\
                11&77.54 & 2.19\\
                12&72.44 & 39.36\\
                13&79.93 & -27.26\\
                14&78.39 & 48.85\\
                15&52.63 & 26.64\\
                16&77.76 & 80.79\\ \bottomrule
            \end{tabular}
        \end{minipage}
    \end{tabularx}
\end{table*}

\subsection{Precipitation map dataset}
\label{sec:experiemnts_dataset}

Copernicus Emergency Management Service (CEMS) offers a variety of meteorological data. These datasets are accessible for various periods on a global and regional scale. The meteorological information obtained from CEMS, namely the ERA5 hourly data on single levels \cite{hersbach2020era5}, is used in this work to collect precipitation data from the Euro-Asian Region. The longitudinal and latitudinal boundaries of the selected area are West: $-31^{\circ}$, East: $100^{\circ}$, South: $15^{\circ}$, and North: $82^{\circ}$.

We have collected hourly precipitation maps over seven years, from January 2016 to December 2022. The first 6 years are used for training, and the last year is used for testing. When training the models, first we shuffle the indices of our dataset, then create a training and validation split with a $0.9/0.1$ ratio. 
The training set ($47348$ samples) is used for training the models, and the validation set ($5260$ samples) is used for model selection. After training the models, we chose the one with the lowest validation loss for each model type. These top-performing models are then tested on the test set ($8760$ samples).
Our goal is to study precipitation maps of disjoint regions.
Therefore, we have selected sixteen disjoint regions on the map, and used their precipitation data. Fig. \ref{fig:map_to_graph} shows the selected sixteen regions placed on the map. These regions form the nodes of the graph and are referred to as R1, R2, $\ldots$, R16, and their positions on the map are tabulated in Table \ref{Tab:region_coords}, and a summary statistics for the training and testing datasets are detailed in Table \ref{tab:statistics}.

\begin{table*}[ht]
\centering
\caption{Summary of key statistics for training and testing Sets.}
\small
\begin{tabular}{l|ll }
    \toprule
   Metrics & Training set & Testing set\\
    \hline
    Grid size & (40, 40) & (40, 40)\\
    Number of samples & 52608 & 8760\\
    Number of regions & 16 & 16 \\
    Min & 0 & 0\\
    Max & 0.0433 & 0.0693\\
    Mean & 7.8359e-5 & 7.5734e-05\\
    Median & 6.5747e-7 & 5.3951e-07\\
    Standard Deviation & 50.000279 & 0.000273\\
    Variance& 7.82239e-8 & 7.4819e-08\\
    Percentage of pixels with rainfall (over 0.0005 mm/h)& 0.04171 & 0.04003\\
    Percentage of pixels with heavy rain (over 0.004 mm/h)& 0.00045 & 0.00041\\
    \bottomrule
\end{tabular}
  \label{tab:statistics}
\end{table*}

\subsection{Studied scenarios}
\label{sec:change_graph_size}

We performed a series of experiments to quantify the nowcasting performance of the proposed GD-CAF model, comparing it with the SmaAt-UNet, RainNet and persistence models. Specifically, we examined three scenarios, which are detailed as follows:

\begin{itemize}
\item \textbf{Ablation study}:

To explore the enhanced impact of pooling within GD-CAF, we assessed the model's performance by testing the inclusion of pooling layers after the input data and/or within the ST-Attention blocks. The detailed cases studied are described in Table \ref{tab:model_types_table}.

\item \textbf{Changing graph size}:

To explore the additional benefits of incorporating multiple disjoint regions, we assessed the models' performance across varying graph sizes. While the models produced predictions for all nodes during training, the testing phase was conducted in two distinct cases: one where Mean Squared Error (MSE) and other metrics were computed exclusively for the target region R1, and another where we evaluated all regions collectively. To attain this objective, we generated five graphs of varying sizes: 1, 2, 4, 8, and 16. In the case of a graph size of 1, it exclusively comprises the region R1. For a graph size of 2, it encompasses both R1 and R2. Similarly, a graph size of 4 includes regions R1 through R4. As we progressively increase the graph size to 8, it spans from R1 to R8. Finally, with a graph size of 16, it encompasses an extensive range, including regions R1 through R16. Our predictions were centered on the upcoming 6 hours, leveraging data from the preceding 6 hours.

 \begin{table*}[ht]
    \centering
    \renewcommand{\tabcolsep}{0pt}
    \renewcommand{\arraystretch}{1}
    \begin{tabularx}{\textwidth}{XXX}
        \begin{minipage}[c]{0.45\textwidth}
            \centering
            \includegraphics[width=\linewidth]{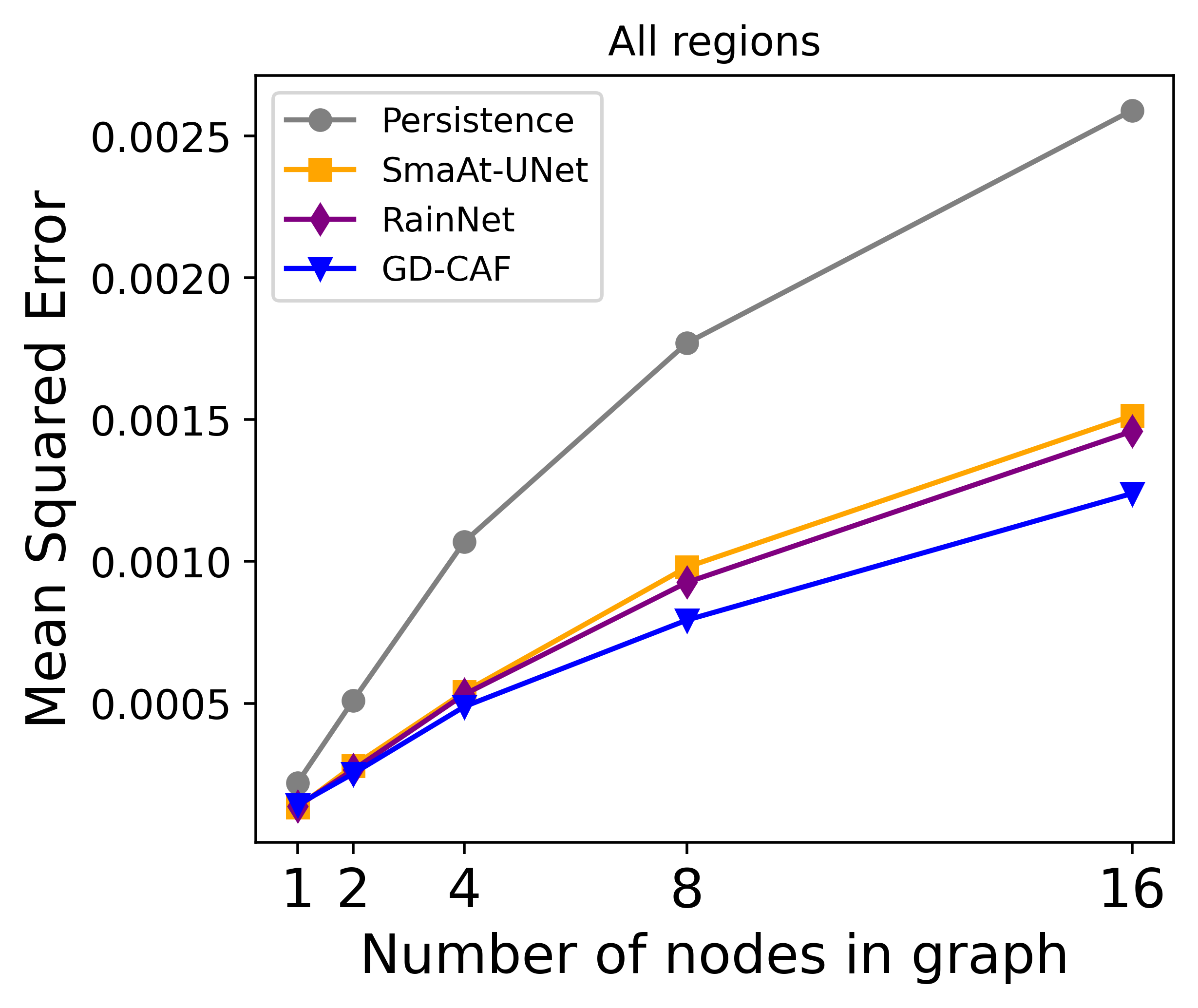}
            \captionof{figure}{Changing the number of nodes in the graph, and calculating MSE on all regions.}
            \label{fig:change_graph_size_all_node}
        \end{minipage}
        &
        \begin{minipage}[c]{0.45\textwidth}
            \centering
            \includegraphics[width=\linewidth]{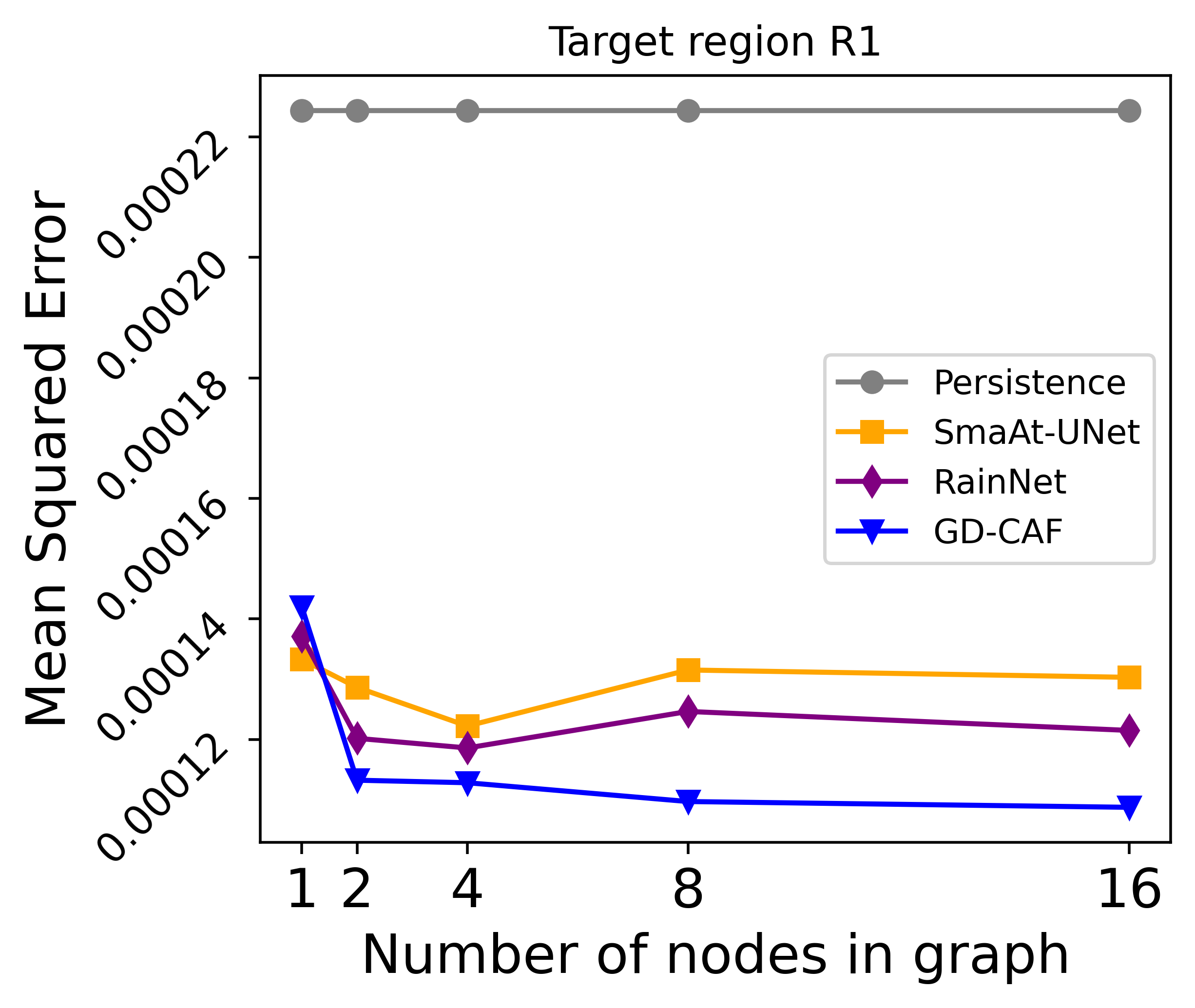}
            \captionof{figure}{Changing the number of nodes in the graph, and calculating MSE only on one target region (R1).} 
            \label{fig:first_node_mse}
        \end{minipage}
    \end{tabularx}
\end{table*}

\item \textbf{Changing number of past observations and prediction time}:

We investigated the performance of the models under variations in both input sample size and prediction time. Specifically, we examined four different input durations, i.e., 6 hours, 9 hours, 12 hours, and 15 hours, and conducted a series of nowcasting tasks for each input size, predicting outcomes at 1-hour, 3-hour, and 6-hour ahead.

\end{itemize}

\section{Results and discussion}
\label{sec:results:discussion}

\subsection{Ablation study}
\label{subsec:effect_each_component}

Unlike the UNet architecture, which commonly reduces input size through downsampling, our model has the tendency to increase input size by incorporating multiple attention heads. This behavior is akin to graph attention models.
To alleviate this computational overhead, we implement pooling techniques to counteract the size increase and uphold computational efficiency.

The performance evaluation of different GD-CAF variants is depicted in Table \ref{tab:model_types_table}. Specifically, GD-CAF (case id=1) exhibits the lowest MSE, while GD-CAF (case id=7) emerges as the most efficient option. In the upcoming experiments, we will utilize the latter variant (case id=7) due to its comparable computational time with SmaAt-UNet, while still maintaining an MSE similar to other GD-CAF variants. Integrating pooling into either the input layer or the computation of query ($Q$), key ($K$), and value ($V$) matrices leads to a slight reduction in performance, albeit still surpassing benchmark models. However, this integration enhances computational speed by 2.13 ($=1920/900$) when pooling is applied to $Q$, $K$, $V$, or by 3.33 $(=1920/576)$ when directly utilizing pooling on the input. Incorporating spatial and temporal attention mechanisms into the model led to a lower mean squared error (MSE) compared to the variant that did not utilize these mechanisms, highlighting that the attention mechanisms improved the model's accuracy.

\begin{table*}[ht]
\caption{An ablation study on the effects of pooling in GD-CAF with regards to MSE. Computational time is the number of seconds one training epoch took, and gain represents the percentage increase in MSE compared to the persistence model.}
\begin{adjustbox}{width=1\textwidth}
\small
\begin{tabular}{ lcccccccccc }
     \toprule
    Model & Case id & \makecell{Pooling \\ Q, K and V}  & \makecell{Pooling \\ Input} & Spatial & Temporal & \makecell{Training Computational\\Time (s/epoch)} & Complexity ↓ & MSE ↓ & \makecell{Gain ↑ \\ (\%)} \\
    \hline
    Persistence & & & & & & 0 & 0 & 0.00259167 & -\\
    SmaAt-UNet & & & & & & 390 & 21.0 M & 0.00151368 & 1.712\\
    RainNet &  &  &  &  &  & 307  & 17.7 M  & 0.00145872  & 1.777 \\
    GD-CAF & 1 &  &  & \checkmark & \checkmark & 1920 & 61.1 K & \underline{0.00122645} & \underline{2.113}\\
    GD-CAF & 2 & \checkmark & & \checkmark & \checkmark & 900 & 70.4 K & 0.00122928 & 2.108\\
    GD-CAF & 3 &  & \checkmark & \checkmark & \checkmark & 576 & 58.2 K & 0.00123289 & 2.102\\
    GD-CAF & 4 & \checkmark & \checkmark & & & \underline{231} & \underline{21.2 K} & 0.00129598 & 2.000\\
    GD-CAF & 5 & \checkmark & \checkmark & \checkmark & & 243 & 42.7 K & 0.00125307 & 2.068\\
    GD-CAF & 6 & \checkmark & \checkmark & & \checkmark & 245 & 42.7 K & 0.00129069 & 2.008\\
    GD-CAF & 7 & \checkmark & \checkmark & \checkmark & \checkmark & 312 & 67.5 K & 0.00123959 & 2.091\\
    \bottomrule
\end{tabular}
\end{adjustbox}
  \label{tab:model_types_table}
\end{table*}

\begin{table*}
  \caption{Test all three models and change the number of nodes in the graph. However, MSE and other metrics were calculated on target region: R1. A ↑ indicates that higher values for that score are better whereas a ↓ indicates that lower scores are better.}
  \centering
\begin{adjustbox}{width=1\textwidth}
\small
\begin{tabular}{clrrrrrrrr}
  \toprule
Nr. nodes & Model & MSE ↓ & Accuracy ↑ & Precision ↑ & Recall ↑ & F1 ↑ & CSI ↑ & FAR ↓ & HSS ↑ \\
\hline
  \multirow{4}{*}{1} & Persistence & 0.00022440 & 0.92906 & 0.23448 & \underline{0.23437} & \underline{0.23442} & \underline{0.13277} & \underline{0.76552} & \underline{0.09861}\\ 
  & SmaAt-UNet & \underline{0.00013329} & \underline{0.95643} & 0.15328 & 0.00021 & 0.00041 & 0.00021 & 0.84672 & 0.00015\\
  & RainNet & 0.00013713  & 0.95365  & 0.49296  & 0.00022  & 0.00043  & 0.00022  & 0.50704  & 0.00020 \\
  & GD-CAF & 0.00014201 & 0.95366 & \underline{0.88889} & 0.00004 & 0.00007 & 0.00004 & \underline{0.11111} & 0.00004\\
  \hline
  \multirow{4}{*}{2} & Persistence & 0.00022440 & 0.92906 & 0.23448 & 0.23437 & 0.23442 & 0.13277 & 0.76552 & 0.09861\\
  & SmaAt-UNet & 0.00012864 & 0.94740 & 0.38652 & 0.22994 & 0.28835 & 0.16846 & 0.61348 & 0.13143\\
  & RainNet & 0.00012019  & 0.95112  & 0.44500  & 0.22106  & 0.29539  & 0.17329  & 0.55500  & 0.13651 \\
  & GD-CAF & \underline{0.00011324} & \underline{0.95381} & \underline{0.50362} & \underline{0.24009} & \underline{0.32517} & \underline{0.19415} & \underline{0.49638} & \underline{0.15217}\\
  \hline
  \multirow{4}{*}{4} & Persistence & 0.00022440 & 0.92906 & 0.23448 & 0.23437 & 0.23442 & 0.13277 & 0.76552 & 0.09861\\ 
  & SmaAt-UNet & 0.00012223 & 0.95227 & 0.45983 & 0.17133 & 0.24964 & 0.14262 & 0.54017 & 0.11514\\
  & RainNet & 0.00011859  & 0.95254  & 0.47150  & 0.19849  & 0.27937  & 0.16237  & 0.52850  & 0.12951 \\
  & GD-CAF & \underline{0.00011282} & \underline{0.95309} & \underline{0.48886} & \underline{0.26972} & \underline{0.34764} & \underline{0.21039} & \underline{0.51114} & \underline{0.16270}\\
  \hline
  \multirow{4}{*}{8} & Persistence & 0.00022440 & 0.92906 & 0.23448 & 0.23437 & 0.23442 & 0.13277 & 0.76552 & 0.09861\\ 
  & SmaAt-UNet & 0.00013153 & 0.94973 & 0.40385 & 0.17796 & 0.24705 & 0.14094 & 0.59615 & 0.11254\\
  & RainNet & 0.00012465  & 0.95093  & 0.41617  & 0.14588  & 0.21604  & 0.12110  & 0.58383  & 0.09836 \\ 
  & GD-CAF & \underline{0.00010970} & \underline{0.95531} & \underline{0.53641} & \underline{0.26345} & \underline{0.35336} & \underline{0.21459} & \underline{0.46359} & \underline{0.16650}\\
  \hline
  \multirow{4}{*}{16} & Persistence & 0.00022440 & 0.92906 & 0.23448 & 0.23437 & 0.23442 & 0.13277 & 0.76552 & 0.09861\\
  & SmaAt-UNet & 0.00013031 & 0.95098 & 0.39373 & 0.10682 & 0.16805 & 0.09173 & 0.60627 & 0.07563\\
  & RainNet & 0.00012150  & 0.95143  & 0.43989  & 0.17565  & 0.25106  & 0.14355  & 0.56011  & 0.11535 \\
  & GD-CAF & \underline{0.00010874} & \underline{0.95438} & \underline{0.51513} & \underline{0.26706} & \underline{0.35176} & \underline{0.21341} & \underline{0.48487} & \underline{0.16529}\\
\bottomrule
\end{tabular}
\end{adjustbox}
  \label{tab:change_graph_nodes}
\end{table*}

\begin{figure*}
    \centering
    \includegraphics[width=\textwidth]{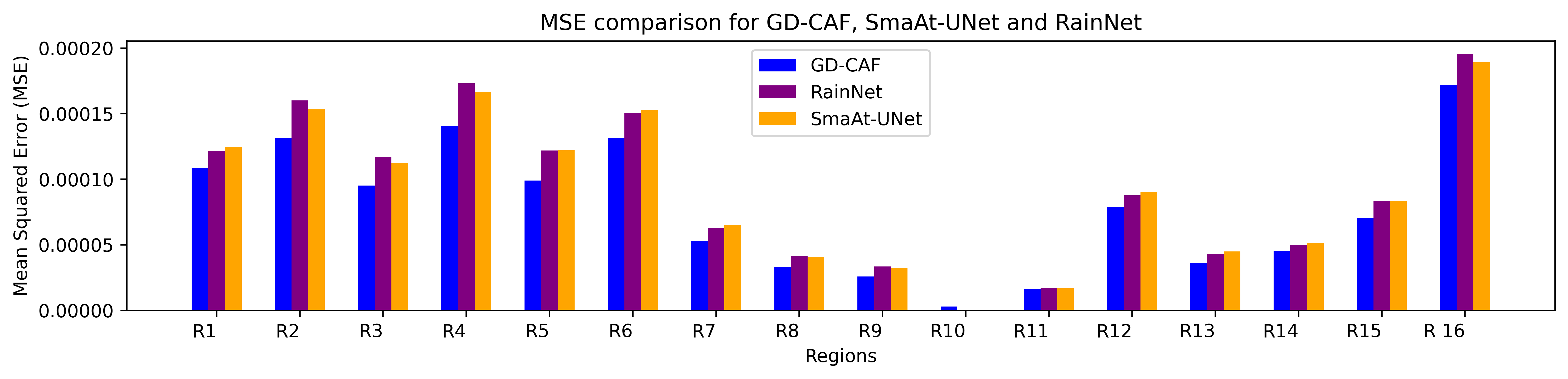}
    \caption{MSE per region for GD-CAF, SmaAt-UNet and RainNet.}
    \label{fig:improvement_every_region}
\end{figure*}

\subsection{Changing graph size}

The obtained overall test MSE values encompassing all regions are shown in Fig. \ref{fig:change_graph_size_all_node}. It can be seen that as the number of regions increases, the overall MSE increases, however, the proposed GD-CAF model surpasses the baseline persistence, RainNet as well as the SmaAt-UNet model.

Fig. \ref{fig:first_node_mse} visualizes the obtained test MSE of only target region $R1$, while the models are trained using different number of nodes. In addition, the obtained results of other tested metrics are shown in Table \ref{tab:change_graph_nodes}. 
When examining a single target region R1, GD-CAF surpasses SmaAt-UNet when the input graph exceeds a size of 1. Notably, the Mean Squared Error (MSE) for SmaAt-UNet gradually decreases as the number of nodes or regions grows, hitting its minimum with 4 input nodes. Yet, post this threshold, the MSE begins to rise. Conversely, for GD-CAF, the MSE continues to decline with an increasing number of nodes, even surpassing four regions. This suggests that GD-CAF effectively learns and leverages the correlations among these distinct regions for enhanced nowcasting. Meanwhile, the performance of the persistence model remains consistent, unaffected by changes in graph size. Additionally, Fig. \ref{fig:improvement_every_region} illustrates the individual improvement achieved for each region when employing GD-CAF compared to SmaAt-UNet. The most significant improvements are observed in $R2$, $R3$, $R4$, and $R6$.

\subsection{Changing input amount and prediction time}

The obtained results of the test set, corresponding to various amounts of past observations utilized by the models to nowcast multiple steps ahead are tabulated in Table \ref{Tab:change_window}. It can be observed that all models outperform the persistence baseline model. However, it's noteworthy that within a 1-hour timeframe, the weather doesn't undergo significant changes at this resolution. Hence, persistence effectively yields accurate predictions for such scenarios.

\begin{figure*}
    \centering
    \includegraphics[width=\textwidth]{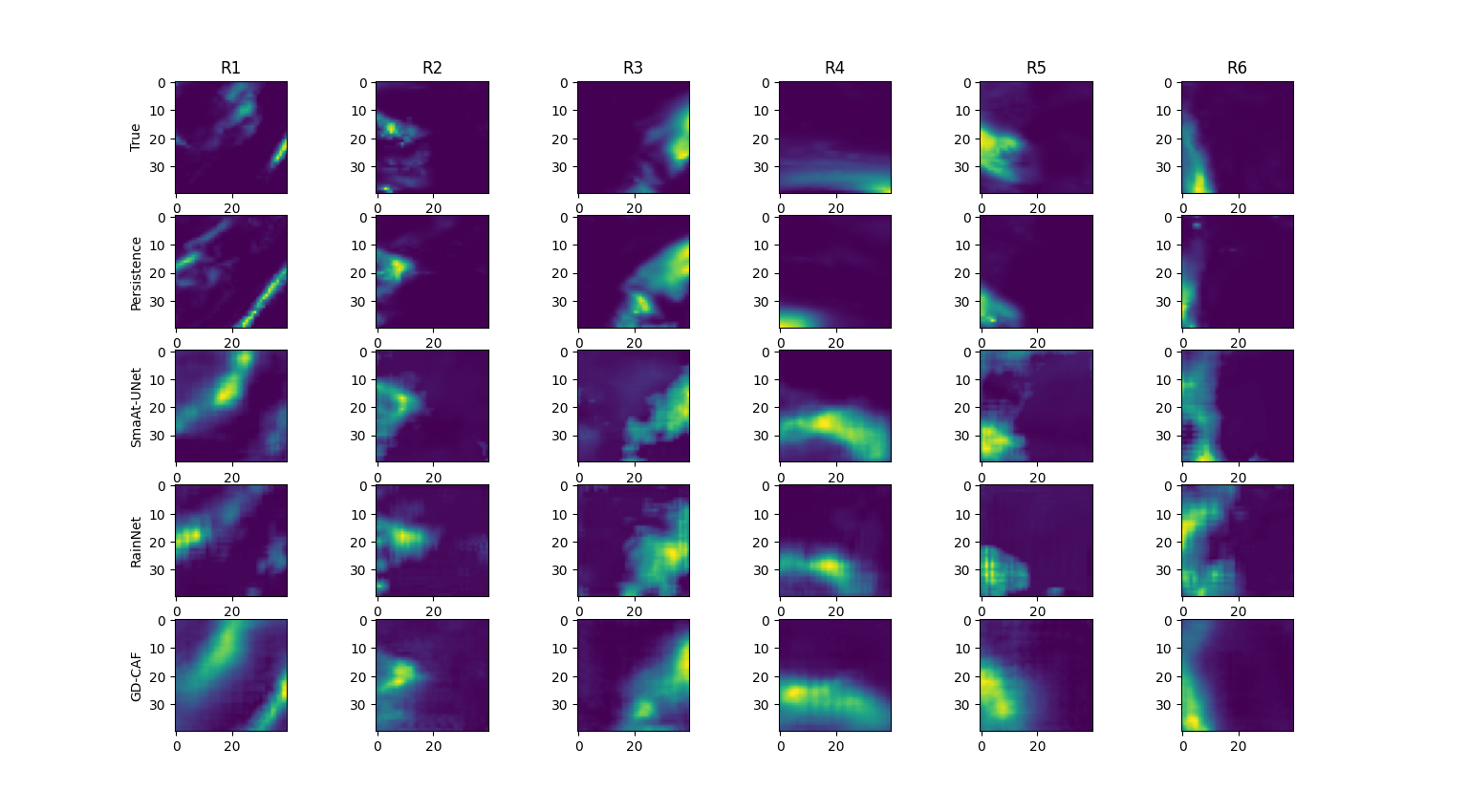}
    \caption{True label, baseline models (Persistence, RainNet, and SmaAt-UNet), GD-CAF prediction from six selected regions.}
    \label{fig:x_y_pred_18}
\end{figure*}

Fig. \ref{fig:avg_mse_on_past_window} shows the test MSE averaged over different input size for all examined models. It can be observed that MSE increases as the future prediction steps increases. Moreover, GD-CAF consistently achieves lower MSE than SmaAt-UNet. Fig. \ref{fig:avg_mse_on_future_window} visualizes the test MSE averaged over prediction steps for all examined models. Furthermore Fig. \ref{fig:x_y_pred_18} shows examples of predictions from various regions.

\begin{table*}
    \centering
    \renewcommand{\tabcolsep}{0pt}
    \renewcommand{\arraystretch}{1}
    \begin{tabularx}{\textwidth}{Xcc}
        \begin{minipage}[c]{0.45\textwidth}
            \centering
            \includegraphics[width=\linewidth]{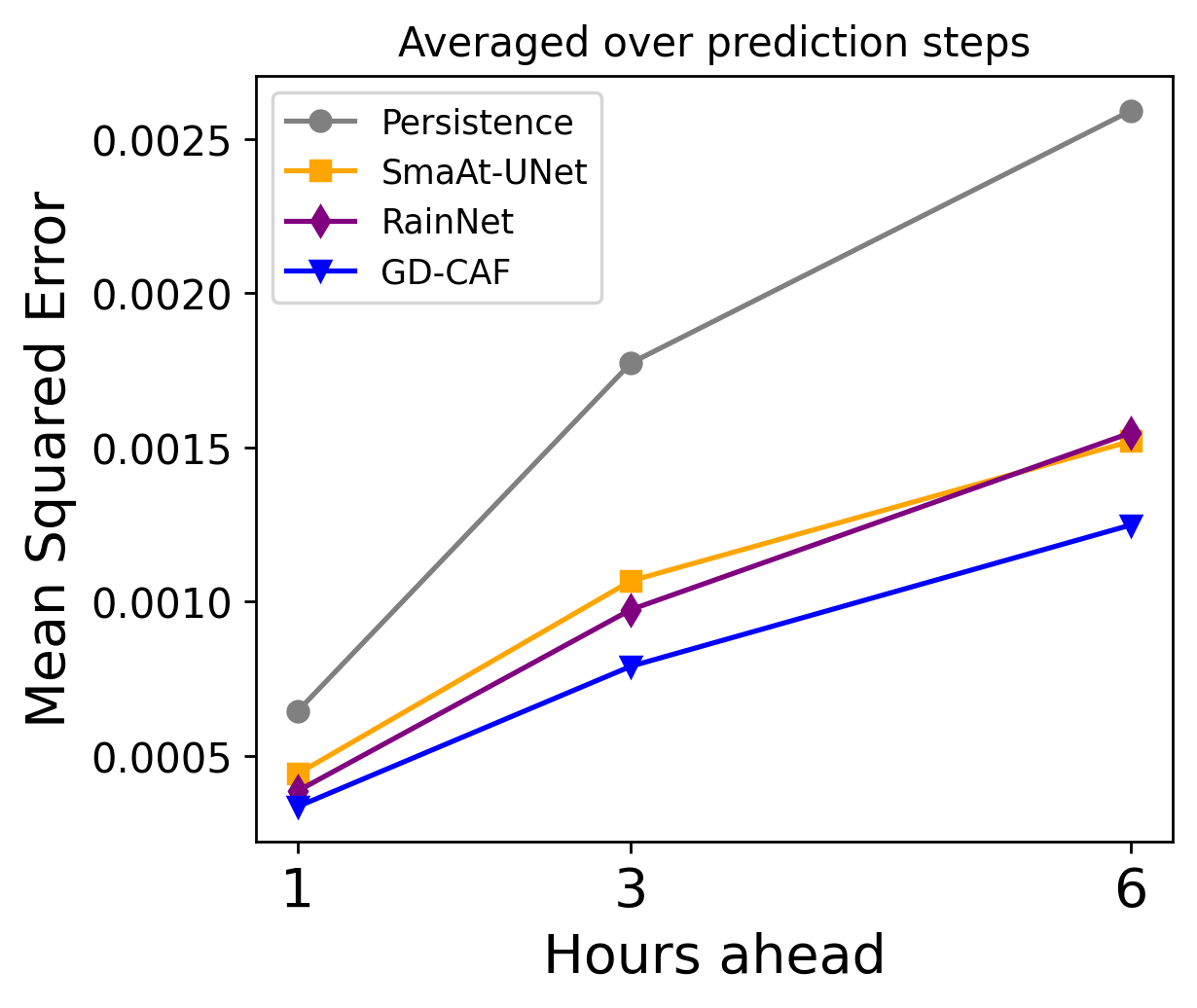}
            \captionof{figure}{MSE of all three models, while changing the past and future window sizes. The performance is averaged on different input window sizes.}
            \label{fig:avg_mse_on_past_window}
        \end{minipage}
        &
        \begin{minipage}[c]{0.45\textwidth}
            \centering
            \includegraphics[width=\linewidth]{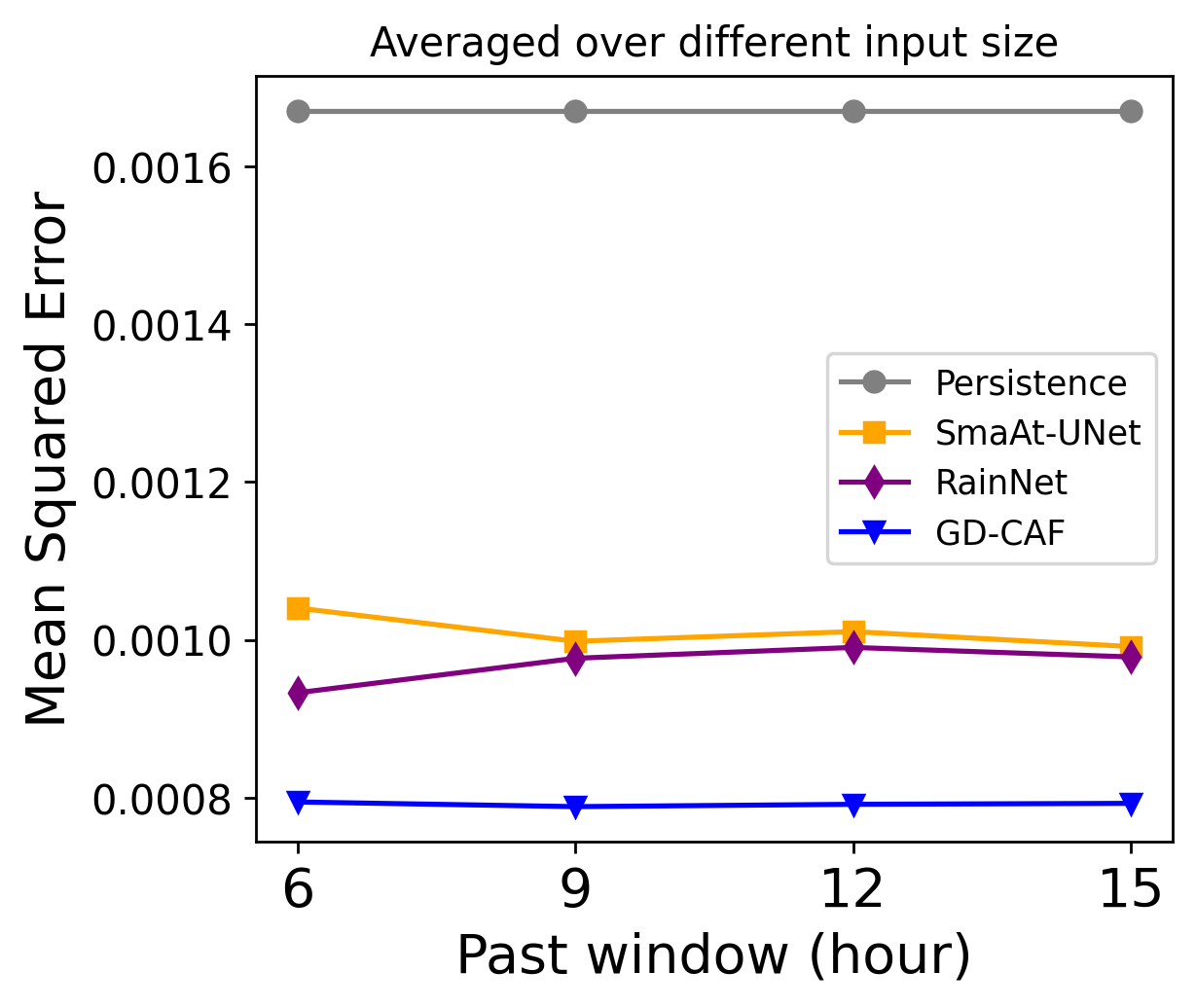}
            \captionof{figure}{MSE of all three models, while changing the past and future window sizes. The performance is averaged on different prediction steps.}
            \label{fig:avg_mse_on_future_window}
        \end{minipage}
    \end{tabularx}
\end{table*}

\begin{figure*}
    \centering
    \includegraphics[width=0.98\textwidth]{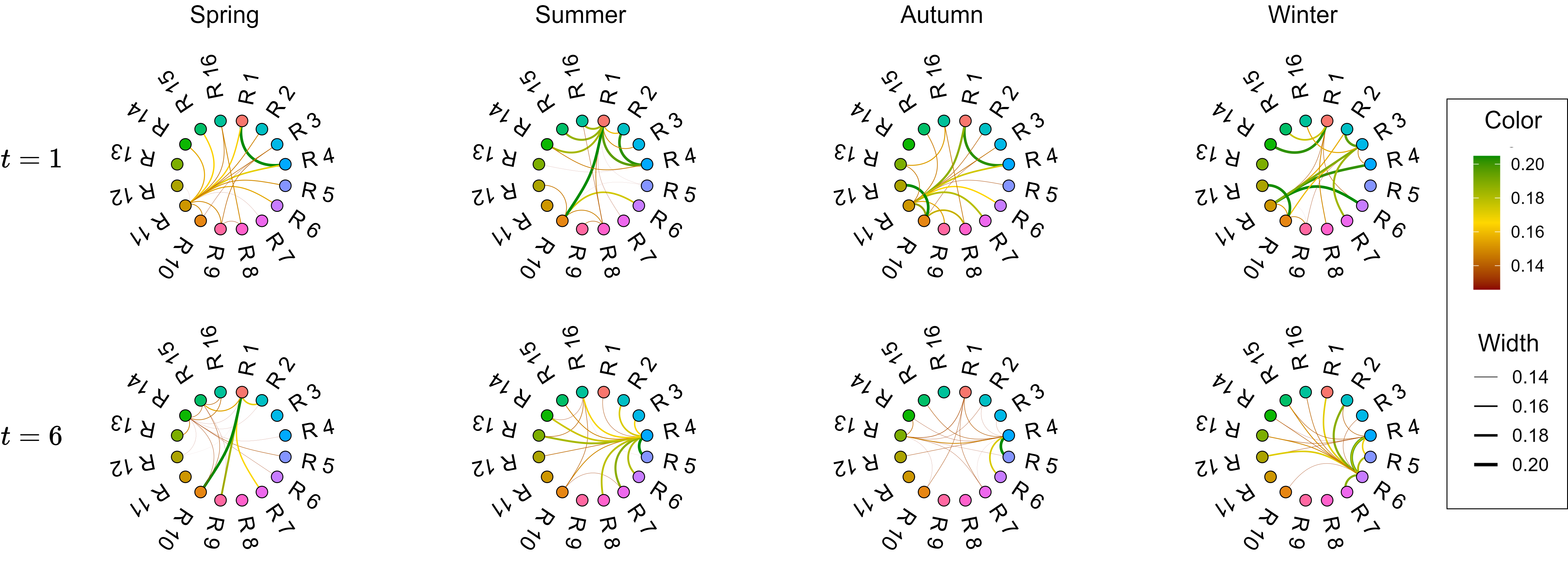}
    \caption{The top 20 strongest spatial attentions are displayed for two time step in a circular graph format. Columns represent seasons, displaying averaged attention matrices across seasons and heads, with prominent correlations observed in $R1$, $R4$, $R6$ and $R11$ with other regions.}
    \label{fig:visualize_attention:spatial}
\end{figure*}

\begin{figure*}
    \centering
    \includegraphics[width=0.98\textwidth]{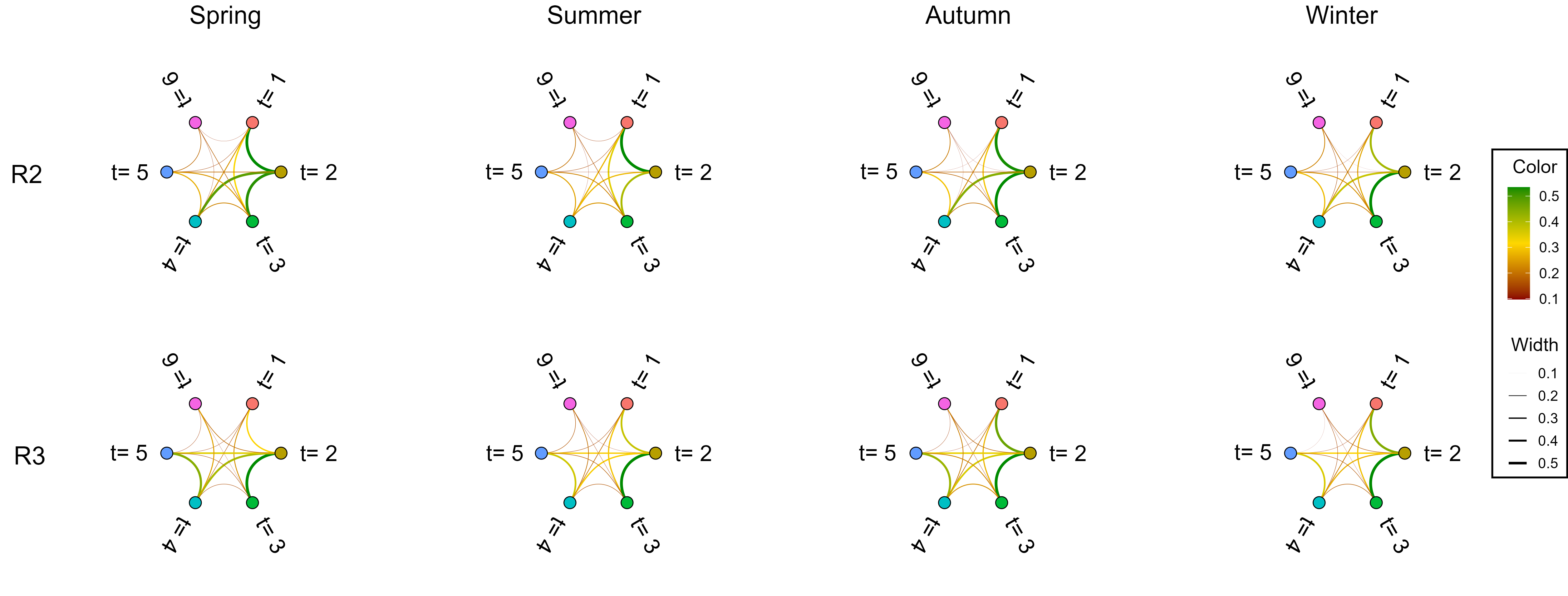}
    \caption{Averaged temporal attentions are displayed for specific regions ($R2, R3$), revealing notable correlations between $t=1$ and $t=2$, and $t=2$ and $t=3$, gradually decreasing afterward. Additionally, correlations between non-consecutive time steps are also observed.}
    \label{fig:visualize_attention:temporal}
\end{figure*}

\begin{table*}
  \caption{Test all three models and change the input amount and prediction time. We used all 16 regions during this experiment. A ↑ indicates that higher values for that score are better whereas a ↓ indicates that lower scores are better.}
  \centering
\begin{adjustbox}{width=1\textwidth}
\small
\begin{tabular}{cclrrrrrrrr}
  \toprule
Input Amount & Prediction & Model & MSE ↓ & Accuracy ↑ & Precision ↑ & Recall ↑ & F1 ↑ & CSI ↑ & FAR ↓ & HSS ↑\\
\hline
   \multirow{12}{*}{6 hour} & \multirow{4}{*}{1 hour} & Persistence & 0.00064499 & 0.97717 & 0.71488 & 0.71486 & 0.71487 & 0.55626 & 0.28512 & 0.35149\\ 
   & & SmaAt-UNet & 0.00041348 & 0.98155 & 0.79804 & 0.72163 & 0.75792 & 0.61020 & 0.20196 & 0.37417\\ 
   & & RainNet & 0.00038051  & 0.98216  & 0.78766  & 0.75884  & 0.77298  & 0.62996  & 0.21234  & 0.38185 \\
   & & GD-CAF & \underline{0.00034831} & \underline{0.98341} & \underline{0.80340} & \underline{0.77519} & \underline{0.78904} & \underline{0.65158} & \underline{0.19660} & \underline{0.39020}\\
   \cline{2-11}
   & \multirow{4}{*}{3 hour} & Persistence & 0.00177373 & 0.95620 & 0.45288 & 0.45285 & 0.45286 & 0.29271 & 0.54712 & 0.21503\\  
   & & SmaAt-UNet & 0.00119339 & 0.96742 & 0.61723 & 0.48978 & 0.54617 & 0.37568 & 0.38277 & 0.26475\\
   & & RainNet & 0.00095934  & 0.96759  & 0.61851  & 0.49648  & 0.55082  & 0.38009  & 0.38149  & 0.26711 \\
   & & GD-CAF & \underline{0.00079560} & \underline{0.97134} & \underline{0.66401} & \underline{0.57462} & \underline{0.61609} & \underline{0.44518} & \underline{0.33599} & \underline{0.30064}\\
   \cline{2-11}
   & \multirow{4}{*}{6 hour} & Persistence & 0.00259167 & 0.94233 & 0.27954 & 0.27952 & 0.27953 & 0.16247 & 0.72046 & 0.12474\\ 
   & & SmaAt-UNet & 0.00151368 & 0.95901 & 0.47165 & 0.20024 & 0.28112 & 0.16355 & 0.52835 & 0.13178\\
   & & RainNet & 0.00145872  & 0.95879  & 0.47151  & 0.24485  & 0.32232  & 0.19212  & 0.52849  & 0.15163 \\
   & & GD-CAF & \underline{0.00123959} & \underline{0.96280} & \underline{0.55945} & \underline{0.33177} & \underline{0.41653} & \underline{0.26305} & \underline{0.44055} & \underline{0.19930}\\
   \hline
   \multirow{12}{*}{9 hour} & \multirow{4}{*}{1 hour} & Persistence & 0.00064499 & 0.97717 & 0.71488 & 0.71486 & 0.71487 & 0.55626 & 0.28512 & 0.35149\\
   & & SmaAt-UNet & 0.00052450 & 0.98096 & 0.77444 & 0.73972 & 0.75668 & 0.60860 & 0.22556 & 0.37339\\
   & & RainNet & 0.00038358  & 0.98223  & 0.79053  & 0.75636  & 0.77307  & 0.63008  & 0.20947  & 0.38191 \\
   & & GD-CAF & \underline{0.00033724} & \underline{0.98385} & \underline{0.81423} & \underline{0.77267} & \underline{0.79291} & \underline{0.65688} & \underline{0.18577} & \underline{0.39226}\\
   \cline{2-11}
   & \multirow{4}{*}{3 hour} & Persistence & 0.00177373 & 0.95620 & 0.45288 & 0.45285 & 0.45286 & 0.29271 & 0.54712 & 0.21503\\ 
   & & SmaAt-UNet & 0.00096567 & 0.96810 & 0.63735 & 0.47065 & 0.54146 & 0.37123 & 0.36265 & 0.26266\\
   & & RainNet & 0.00096393  & 0.96743  & 0.61492  & 0.49815  & 0.55041  & 0.37970  & 0.38508  & 0.26685 \\
   & & GD-CAF & \underline{0.00079051} & \underline{0.97178} & \underline{0.67401} & \underline{0.57117} & \underline{0.61834} & \underline{0.44754} & \underline{0.32599} & \underline{0.30190}\\
   \cline{2-11}
   & \multirow{4}{*}{6 hour} & Persistence & 0.00259167 & 0.94233 & 0.27954 & 0.27952 & 0.27953 & 0.16247 & 0.72046 & 0.12474\\ 
   & & SmaAt-UNet & 0.00150384 & 0.95922 & 0.47699 & 0.19653 & 0.27836 & 0.16169 & 0.52301 & 0.13055\\
   & & RainNet & 0.00158214  & 0.95837  & 0.45596  & 0.20856  & 0.28621  & 0.16700  & 0.54404  & 0.13391 \\
   & & GD-CAF & \underline{0.00123802} & \underline{0.96257} & \underline{0.55102} & \underline{0.34927} & \underline{0.42754} & \underline{0.27189} & \underline{0.44898} & \underline{0.20460}\\ 
   \hline
   \multirow{12}{*}{12 hour} & \multirow{4}{*}{1 hour} & Persistence & 0.00064499 & 0.97717 & 0.71488 & 0.71486 & 0.71487 & 0.55626 & 0.28512 & 0.35149\\ 
   & & SmaAt-UNet & 0.00039728 & 0.98180 & 0.79094 & 0.74110 & 0.76521 & 0.61971 & 0.20906 & 0.37788\\ 
   & & RainNet & 0.00039026  & 0.98199  & 0.78639  & 0.75527  & 0.77052  & 0.62670  & 0.21361  & 0.38058 \\
   & & GD-CAF & \underline{0.00033237} & \underline{0.98385} & \underline{0.80460} & \underline{0.78770} & \underline{0.79606} & \underline{0.66121} & \underline{0.19540} & \underline{0.39383}\\
   \cline{2-11}
   & \multirow{4}{*}{3 hour} & Persistence & 0.00177373 & 0.95620 & 0.45288 & 0.45285 & 0.45286 & 0.29271 & 0.54712 & 0.21503\\
   & & SmaAt-UNet & 0.00110666 & 0.96587 & 0.60281 & 0.43151 & 0.50298 & 0.33599 & 0.39719 & 0.24290\\
   & & RainNet & 0.00102656  & 0.96581  & 0.58746  & 0.48905  & 0.53376  & 0.36403  & 0.41254  & 0.25808 \\
   & & GD-CAF & \underline{0.00078344} & \underline{0.97176} & \underline{0.67074} & \underline{0.57811} & \underline{0.62099} & \underline{0.45032} & \underline{0.32926} & \underline{0.30320}\\
   \cline{2-11}
   & \multirow{4}{*}{6 hour} & Persistence & 0.00259167 & 0.94233 & 0.27954 & 0.27952 & 0.27953 & 0.16247 & 0.72046 & 0.12474\\ 
   & & SmaAt-UNet & 0.00152703 & 0.95789 & 0.44970 & 0.23396 & 0.30779 & 0.18189 & 0.55030 & 0.14415\\
   & & RainNet & 0.00155397  & 0.95752  & 0.43597  & 0.21014  & 0.28359  & 0.16522  & 0.56403  & 0.13222 \\
   & & GD-CAF & \underline{0.00125883} & \underline{0.96258} & \underline{0.55336} & \underline{0.33652} & \underline{0.41852} & \underline{0.26464} & \underline{0.44664} & \underline{0.20019}\\ 
   \hline
   \multirow{12}{*}{15 hour} & \multirow{4}{*}{1 hour} & Persistence & 0.00064499 & 0.97717 & 0.71488 & 0.71486 & 0.71487 & 0.55626 & 0.28512 & 0.35149\\ 
   & & SmaAt-UNet & 0.00043228 & 0.98127 & 0.77626 & 0.74734 & 0.76153 & 0.61489 & 0.22374 & 0.37589\\ 
   & & RainNet & 0.00039041  & 0.98199  & 0.78834  & 0.75193  & 0.76971  & 0.62563  & 0.21166  & 0.38017 \\
   & & GD-CAF & \underline{0.00032502} & \underline{0.98410} & \underline{0.81325} & \underline{0.78223} & \underline{0.79744} & \underline{0.66312} & \underline{0.18675} & \underline{0.39458}\\
   \cline{2-11}
   & \multirow{4}{*}{3 hour} & Persistence & 0.00177373 & 0.95620 & 0.45288 & 0.45285 & 0.45286 & 0.29271 & 0.54712 & 0.21503\\  
   & & SmaAt-UNet & 0.00100494 & 0.96761 & 0.63306 & 0.45344 & 0.52840 & 0.35907 & 0.36694 & 0.25605\\
   & & RainNet & 0.00094797  & 0.96764  & 0.61762  & 0.50207  & 0.55388  & 0.38301  & 0.38238  & 0.26864 \\
   & & GD-CAF & \underline{0.00079395} & \underline{0.97193} & \underline{0.67516} & \underline{0.57532} & \underline{0.62126} & \underline{0.45060} & \underline{0.32484} & \underline{0.30339}\\
 \cline{2-11}
   & \multirow{4}{*}{6 hour} & Persistence & 0.00259167 & 0.94233 & 0.27954 & 0.27952 & 0.27953 & 0.16247 & 0.72046 & 0.12474\\ 
   & & SmaAt-UNet & 0.00153682 & 0.95880 & 0.46486 & 0.19871 & 0.27841 & 0.16172 & 0.53514 & 0.13035\\
   & & RainNet & 0.00159625  & 0.95624  & 0.40711  & 0.20614  & 0.27370  & 0.15854  & 0.59289  & 0.12681 \\
   & & GD-CAF & \underline{0.00125927} & \underline{0.96270} & \underline{0.55536} & \underline{0.33808} & \underline{0.42030} & \underline{0.26606} & \underline{0.44464} & \underline{0.20110}\\
\bottomrule
\end{tabular}
\end{adjustbox}
  \label{Tab:change_window}
\end{table*}

\subsection{Insights and Interpretation}

Fig. \ref{fig:visualize_attention:spatial} displays spatial attentions in the last ST-Attention block in a circular graph format. Each graph shows the top 20 strongest attention values between different regions. The first row contains spatial attention matrices from the first time step, while the second row contains those from the last time step. Each column represents a different season, with the spatial attention matrices averaged for each season and across attention heads. Notably, when $t=1$, $R1$ exhibited a strong correlation with $R4$, $R10$, and $R14$. $R11$ showed a lot of strong connections in spring, autumn, and winter. When $t=6$, we noticed one region in each season that has high cardinality; these are $R14$, $R4$, $R13$, and $R6$. Overall, a strong correlation does not necessarily imply that the cells are physically close to each other. However, in the cases of $R1 - R15$, $R4 - R5$, and $R4 - R6$, this is indeed the case. Fig. \ref{fig:visualize_attention:temporal} presents the temporal attentions for selected regions ($R2, R3$). These temporal attention values were averaged across each season in the test set. Notably, a strong correlation is observed between $t=1$ and $t=2$, as well as between $t=2$ and $t=3$. However, after these time steps, the correlation gradually decreases. It's also worth mentioning that there is some correlation between non-consecutive time steps.

\subsection{Model Strengths and Limitations}

GD-CAF outperforms other models while using fewer parameters. Its integration of spatial and temporal attention mechanisms enhances the model’s ability to learn patterns in regional interactions and temporal dependencies. However, the increased computational complexity due to multiple attention heads can potentially pose challenges in large-scale applications. Additionally, while pooling techniques improve efficiency, they slightly reduce performance, which may affect scenarios requiring maximum accuracy.

\section{Conclusion}
\label{sec:conclusion}

In this study, we introduce an innovative Graph Dual-stream Convolutional Attention Fusion (GD-CAF) model tailored for precipitation nowcasting tasks. The incorporation of spatiotemporal convolutional attention and gated fusion modules, enhanced by depthwise-separable convolutional layers, enables our model to extract valuable insights from high-dimensional spatiotemporal graph nodes that encapsulate historical precipitation maps. Leveraging direct exploration of higher-order correlations among input data dimensions, our approach proves particularly advantageous in scenarios constrained by limited access to local information across multiple regions. Rigorous evaluations against benchmark models such as SmaAt-UNet, RainNet and persistence models attest to the superior performance of our GD-CAF model under varying conditions. Notably, our model surpasses these benchmarks, showcasing its efficacy in handling the complexities of nowcasting tasks. To offer additional insights into the predictive capabilities of our model, we visualize the strongest connections between regions or time periods. This visualization is derived from the averaged spatial and temporal attention scores computed across each season within test set. The implementation of our proposed model, including the trained models, can be found on GitHub at \url{https://github.com/wendig/GD-CAF}.

\section{Bibliography}

\bibliographystyle{elsarticle-num} 
\bibliography{cas-refs}

\end{document}